\documentclass[10pt,twocolumn,letterpaper]{article}

\usepackage{iccv}
\usepackage{times}
\usepackage{epsfig}
\usepackage{graphicx}
\usepackage{amsmath}
\usepackage{amssymb}
\usepackage[title]{appendix}
\usepackage[caption=false]{subfig}
\usepackage{caption}
\usepackage{stfloats}
\usepackage[dvipsnames]{xcolor}
\usepackage[ruled,vlined]{algorithm2e}
\usepackage[pagebackref=true,breaklinks=true,letterpaper=true,colorlinks,bookmarks=false]{hyperref}

\SetKwInput{KwInput}{Input}
\SetKwInput{KwOutput}{Output}

\usepackage[pagebackref=true,breaklinks=true,letterpaper=true,colorlinks,bookmarks=false]{hyperref}

\iccvfinalcopy 

\def\iccvPaperID{7663} 

\ificcvfinal\fi

\begin{document}

\title{HeadGAN: One-shot Neural Head Synthesis and Editing}

\author{Michail Christos Doukas$^{1,2}$, Stefanos Zafeiriou$^{1,2}$, Viktoriia Sharmanska$^{1,3}$\\
$^1$Imperial College London, UK $\quad ^2$Huawei Technologies, London, UK $\quad^3$ University of Sussex, UK\\
{\tt\small $\{$michail-christos.doukas16, s.zafeiriou, sharmanska.v$\}$@imperial.ac.uk}
}

\twocolumn[{%
\renewcommand\twocolumn[1][]{#1}%
\maketitle
\begin{center}
    \begin{picture}(225,225)
    \put(-125,0){\includegraphics[width=0.95\linewidth]{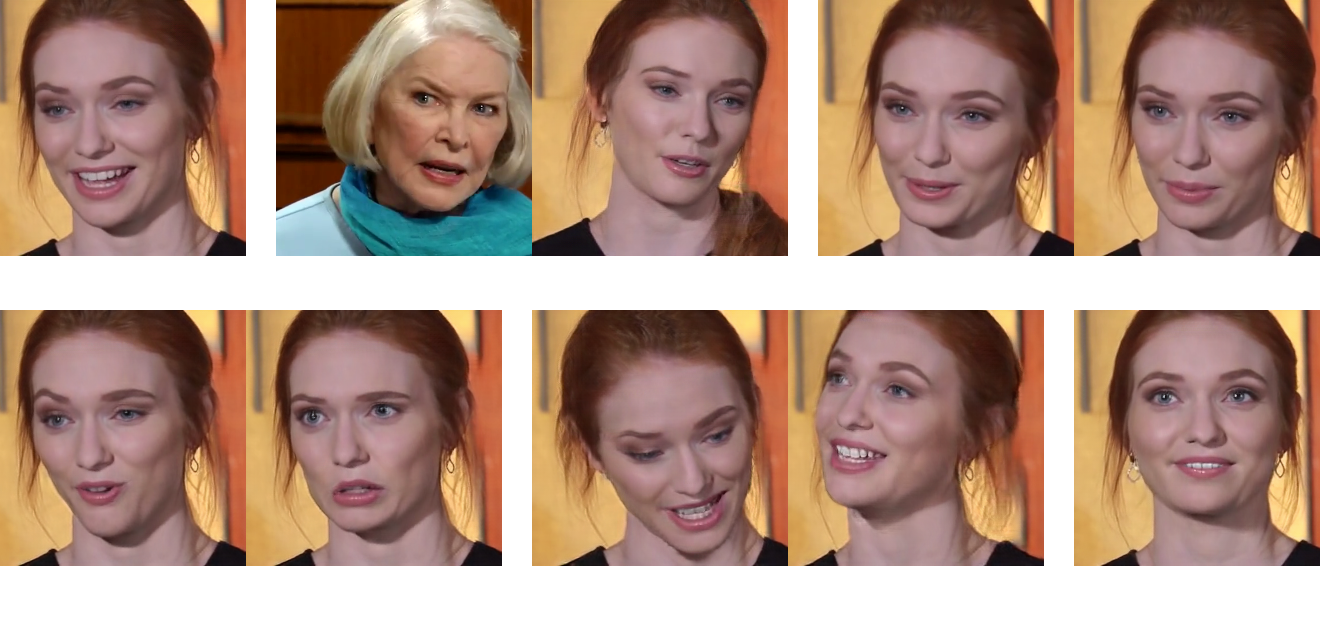}}
    \put(-123,120){(a) Reference image}
    \put(30,120){(b) Reenactment}
    \put(180,120){(c) Reconstruction (self-reenactment)}
    \put(-85,10){(d) Expression Editing}
    \put(118,10){(e) Pose Editing}
    \put(265,10){(f) Frontalisation}
    \end{picture}
    \captionof{figure}{Our proposed \textit{HeadGAN} method performs reenactment (b), by fully transferring the facial expressions and head pose from a driving frame to a reference image. When the driving and reference identities coincide (c), it can be used for facial video compression and reconstruction. In addition, \textit{HeadGAN} can be applied to facial expression editing (d), novel view synthesis (e) and face frontalisation (f). Project's page: \url{https://michaildoukas.github.io/HeadGAN/}}
    \label{fig:fig1}
\end{center}%
}]


\ificcvfinal\thispagestyle{empty}\fi

\begin{abstract}
Recent attempts to solve the problem of head reenactment using a single reference image have shown promising results. However, most of them either perform poorly in terms of photo-realism, or fail to meet the identity preservation problem, or do not fully transfer the driving pose and expression. We propose HeadGAN, a novel system that conditions synthesis on 3D face representations, which can be extracted from any driving video and adapted to the facial geometry of any reference image, disentangling identity from expression. We further improve mouth movements, by utilising audio features as a complementary input. The 3D face representation enables HeadGAN to be further used as an efficient method for compression and reconstruction and a tool for expression and pose editing.
\end{abstract}

\section{Introduction}

Visual data synthesis \cite{wang2018vid2vid, fsvid2vid}, including talking head animation \cite{X2Face, Zakharov2019FewShotAL, bilayer, MarioNETte:AAAI2020, Siarohin_2019_CVPR, FOMM, kim2018deep} are particularly exciting and thriving research areas, with countless applications in editing, games, social media, VR, teleconference and virtual assistance. Over the past years, solutions were mainly given by the graphics community. For instance, \textit{Face2Face}~\cite{face2face} method performs face reenactment, by recovering facial expressions from a driving video and overwriting them to the source frames. Some recent learning-based approaches \cite{kim2018deep, head2head2020, head2headpp} have sought to solve the problem of full head reenactment, which aims to transfer not only the expression, but also the pose, from a driving person to the source identity. The shortcoming of such methods is their dependence on long video footage of the source, as they train person-specific models. At the same time, various methods have been proposed for reenacting human heads under a few-shot setting \cite{X2Face, fsvid2vid, bilayer, MarioNETte:AAAI2020, FOMM, Geng2018WarpguidedGF}, where only a limited number of reference images are available, even a single one. Most state-of-the-art approaches use facial key-points to guide synthesis \cite{Zakharov2019FewShotAL, bilayer, fsvid2vid, FOMM}, which usually leads to \textit{identity preservation problems} during reenactment, as key-points encode appearance information. The problem becomes more prominent when the head geometry of the source differs from that of the person in the driving video.

In this paper we propose \textit{HeadGAN}, a novel one-shot GAN-based method for head animation and editing. We take a different approach from most existing few-shot methods and use \textit{a 3D face representation} \textcolor{black}{similar to PNCC \cite{PNCC}} to condition synthesis. We capitalise on prior knowledge of expression and identity disentanglement, enclosed within \textit{3D Morphable Models (3DMMs)} \cite{3dmms, 3dmmsbooth, Booth, Booth_2016_CVPR}. Our decision to model faces with 3DMMs enables \textit{HeadGAN} to operate as: 1) a real-time \textit{reenactment system} \textcolor{black}{operating at $\sim$ 20 fps}, 2) an efficient method for facial video \textit{compression and reconstruction}, 3) a facial \textit{expression editing method}, 4) a \textit{novel view synthesis system}, including face \textit{frontalisation}. Fig. 1 illustrates the tasks supported by our method. Apart from 3D faces, we \textcolor{black}{optionally} condition the generative process on speech features coming from the audio signal, enabling our method to perform accurate mouth synthesis, as suggested by our automated lipreading experiment.

We perform extensive comparisons with state-of-the-art methods \cite{X2Face, fsvid2vid, bilayer, FOMM, psp, RaR} and report superior image quality and performance, in terms of standard GAN metrics \cite{NIPS2017_8a1d6947, unterthiner2018towards}, on the tasks of reconstruction, reenactment and frontalisation, even when compared to models \cite{bilayer} trained on the larger VoxCeleb2 \cite{Chung18b} dataset. Lastly, we conduct an ablation study in order to demonstrate the contribution of each component of our system.

\section{Related Work}

\noindent \textbf{Model-free methods for face synthesis.} \textit{X2Face} \cite{X2Face} is among the earliest learning-based methods for animating human heads that does not rely on any prior knowledge of faces. In some cases their warping operation causes unnatural head deformations, leading to poor photo-realism. \textit{MonkeyNet} \cite{Siarohin_2019_CVPR} is a more recent deep learning framework that proposes to infer motion via key-point detection from the driving video. Then, the appearance extracted from the reference image along with motion information are used to generate the output. In the follow-up work, \textit{First Order Motion Model (FOMM)} \cite{FOMM} significantly improves the results of single image animation. \textit{FOMM} uses relative key-point locations in order to preserve the identity of the source, which requires the object in the first frame of the driving video to be in the same pose with the one in the source image. As such assumption is not always met, the head pose of generated samples is not guaranteed to follow the driver.

\noindent \textbf{Landmark-based face modeling and generation.} \textit{Bringing Portraits to Life} \cite{elor2017bringingPortraits} is one of the initial attempts to animate still images. 2D warps are applied on the source image in order to imitate the facial transformations in the driving video. It shows promising results when the source head pose is close to the one appearing in the target image and only a small deformation is required. \textit{Warp-Guided GANs}~\cite{Geng2018WarpguidedGF} is a more recent work that uses 2D facial landmarks and 2D warps to animate an image. It requires a photo captured in a frontal pose with neutral expression. Much research on head animation assumes a few-show setting, where a small number of reference images are available. Zakharov \etal \cite{Zakharov2019FewShotAL} extracts identity related embeddings from the reference images and injects them into the generator through adaptive instance normalisation layers (AdaIN) \cite{adain}. Their image-based method performs best after fine-tuning on the new identity. \textit{Bi-layer Neural Avatars}~\cite{bilayer} is another one-shot model that capitalises on SPADE \cite{park2019SPADE}, while operating in real-time speeds during inference. Using SPADE layers \cite{park2019SPADE} for the adaptation of the generative process to the appearance of the source has been proposed earlier, as part of the \textit{few-shot vid2vid} model~\cite{fsvid2vid}, which is a video-based method extending \textit{vid2vid}~\cite{wang2018vid2vid}. Most aforementioned methods are not able to address the identity preservation problem in reenactment, since facial landmarks allow identity related information from the driver to be transferred into the generated samples. \textit{MarioNETte}~\cite{MarioNETte:AAAI2020} tries to solve this problem, by proposing a method for landmark transformation that adapts the driving landmarks to the reference head shape.

\noindent \textbf{Head animation assisted by 3D faces.} 3DMMs \cite{3dmms, 3dmmsbooth, Booth, Booth_2016_CVPR} have been proven to be very effective for modeling human faces and have been widely used to drive face synthesis \cite{face2face, kim2018deep, ijcai2019-129, thies2020nvp}. Fitting 3DMMs, enables recovering accurate pose and expressions from the target frames, as well as identity related parameters from the reference image(s). Then, the rendered 3D faces are used to condition neural networks, which complete the texture and fill in the areas of missing information (hair, body, background, etc.). \textit{Deep video portraits (DVP)} \cite{kim2018deep} and \textit{Head2Head} \cite{head2head2020} are examples of such reenactment systems driven by 3D information. Both methods train \emph{person specific} models, using a long video footage of the source subject. On the contrary, our proposed approach is \emph{person generic}, \textit{i.e.}~it can perform video synthesis for any unseen person, using a single reference image. \textcolor{black}{Other methods such as \textit{StyleRig} \cite{stylerig} and \textit{GIF} \cite{GIF2020} use 3DMMs to control \textit{StyleGAN} and \textit{StyleGAN2} \cite{stylegan, stylegan2}, but fail to  preserve parts of the scene that are not explained by the face models, such as hair, background.}

\noindent \textbf{Audio-driven head synthesis.} Apart from the video-driven techniques discussed above, there exists an extensive body of literature focusing on audio-driven talking face synthesis \cite{obama, Chung17b, ijcai2019-129, chen2019hierarchical, Vougioukas, thies2020nvp}. Much different from these methods, our system \textcolor{black}{can optionally use} audio signals to enhance speech quality and realism within the mouth area, while the pose and expressions are guided by the target video footage.

\section{Methodology}

\subsection{3D Face Representation}
\label{sec:face_representation}

In order to accurately transfer the expressions of the driving person while preserving the facial geometry of the source identity, we take advantage of prior knowledge of human faces, contained within 3DMMs \cite{3dmms, 3dmmsbooth, Booth, Booth_2016_CVPR}. Given a driving video of $T$ frames, $\textbf{y}_{1:T} = \{\textbf{y}_t \mid {t=1,\dots,T}\}$, the 3DMM fitting stage produces a sequence of \textit{camera parameters} $\textbf{c}_{1:T}$ and \textit{shape parameters} $\textbf{p}_{1:T}$, with $\textbf{p}_t=[ \textbf{p}_t^{id\top} ; \textbf{p}_t^{exp\top}]^{\top}$. That is, for each frame $t$, we obtain two types of shape parameters: a) identity related parameters $\textbf{p}_t^{id} \in {\rm I\!R}^{n_{id}}$, encoding facial geometry and b) expression parameters $\textbf{p}_t^{exp} \in {\rm I\!R}^{n_{exp}}$, representing facial deformations. This enables disentangling facial shape attributes that depend on identity from shape deformations caused by motion.  We recover very accurate facial expressions, as our 3DMM fitting stage relies on a dense set of 3D points (around 1K). These points are regressed from frames with \textit{RetinaFace} \cite{RetinaFace}, \textcolor{black}{which is pre-trained on WIDER FACE dataset \cite{yang2016wider}}.
Additionally, given a reference image of the source identity $\textbf{y}_{ref}$, we perform 3DMM fitting to obtain the source's shape parameters $\textbf{p}_{ref}^{id}$, $\textbf{p}_{ref}^{exp}$ and camera parameters $\textbf{c}_{ref}$. For details on the 3DMM fitting algorithm please refer to Appendix~\ref{appendix:A}.

\begin{figure}[t!]
\centering
\includegraphics[scale=0.58]{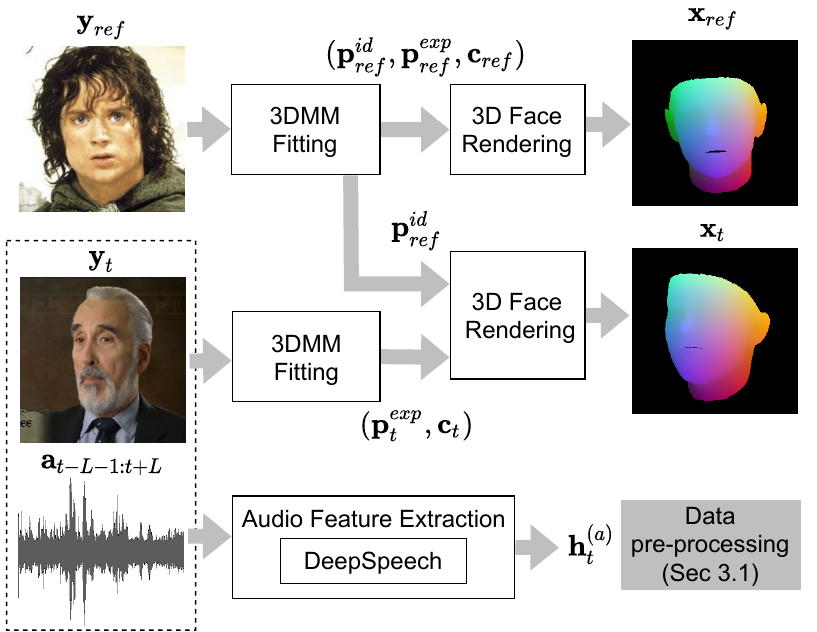}
\caption{Data pre-processing stage. We recover and render the 3D face of the reference image $\textbf{y}_{ref}$, as well as the driving frame $\textbf{y}_t$, after adapting the identity parameters.}
\label{fig:fig2}
\end{figure}

Next, for each frame $t$, we compute the 3D facial shape (3D mesh) $\textbf{s}_t=[x_1, y_1, z_1,..., x_N, y_N, z_N]^{\top} \in {\rm I\!R}^{3N}$, as
\begin{equation}
\label{eq:3DMM}
\textbf{s}_t = \bar{\textbf{x}} + \textbf{U}^{id} \textbf{p}_{ref}^{id} + \textbf{U}^{exp} \textbf{p}_t^{exp}.
\end{equation}
Here $\bar{\textbf{x}} \in {\rm I\!R}^{3N}$ is the mean shape, $\textbf{U}_{id}$ is the identity orthonormal basis and $\textbf{U}_{exp}$ is the expression orthonormal basis of LSFM morphable model \cite{Booth_2016_CVPR}. By construction, this 3D shape $\textbf{s}_t$ reflects the facial structure $\textbf{p}_{ref}^{id}$ of the source  with the facial expressions $\textbf{p}_t^{exp}$ of the driving identity. In this way, we address the source identity preservation problem.
Finally, we render a 3D face representation $\textbf{x}_t = \mathcal{R}(\textbf{s}_t, \textbf{c}_t)$, 
using the 3D shape $\textbf{s}_t$ and camera parameters $\textbf{c}_t$. This is an RGB image \textcolor{black}{similar to PNCC \cite{PNCC}}, as shown in Fig. \ref{fig:fig2}. We also render $\textbf{x}_{ref}$, which is the 3D face recovered from the reference image $\textbf{y}_{ref}$, using camera parameters $\textbf{c}_{ref}$ and shape $\textbf{s}_{ref}$ that is obtained from $\textbf{p}_{ref}^{id}$ and $\textbf{p}_{ref}^{exp}$ using Eq.~\ref{eq:3DMM}. \textcolor{black}{In Sec.~\ref{sec:details} we discuss 3D Face Rendering in more detail.}

Summarising, given a driving video $\textbf{y}_{1:T}$ and a source image $\textbf{y}_{ref}$, the data pre-processing pipeline recovers a sequence of images $\textbf{x}_{1:T}$, which depict the 3D face extracted from the driver and adapted to the facial geometry of the source, as well as the 3D face of the reference image $\textbf{x}_{ref}$. These facial representations are used to condition image synthesis with \textit{HeadGAN}'s Generator network.

\subsection{Audio features}

As opposed to previous one-shot head reenactment systems, our method takes advantage of the driving audio stream and its correlation with facial and mouth movements. We split the audio signal into $T$ parts $\textbf{a}_{1:T}$, where each part $\textbf{a}_{t}$ is aligned and corresponds to frame $\textbf{y}_{t}$ of the driving video with length $T$. Then, we apply audio feature extraction to a window of $2L$ audio parts $\textbf{a}_{t-L-1:t+L} = \{ \textbf{a}_{t-L-1}, \dots, \textbf{a}_{t}, \dots, \textbf{a}_{t+L} \}$, centred around frame $t$, to obtain a feature vector $\textbf{h}_{t}^{(a)}$, which contains information from the past and future time steps. We employ \cite{giannakopoulos2015pyaudioanalysis} for the extraction of low level features, such as MFCCs, signal energy and entropy, which yields a feature vector $\textbf{h}_{t}^{(a_L)} \in {\rm I\!R}^{84}$. Then, we use DeepSpeech \cite{deepspeech} for the extraction of character level logits from each part $\textbf{a}_{t'} \in \textbf{a}_{t-L-1:t+L}$. This results in $2L$ logits, which after concatenation gives a feature vector $\textbf{h}_{t}^{(a_H)} \in {\rm I\!R}^{2L \cdot 27}$. Our final audio feature vector is given as $\textbf{h}_{t}^{(a)} = [\textbf{h}_{t}^{{(a_L)}^{\top}}; \textbf{h}_{t}^{{(a_H)}^{\top}}]^{\top} \in {\rm I\!R}^{300}$, for $L=4$.

\begin{figure*}[t!]
\centering
\includegraphics[scale=0.226]{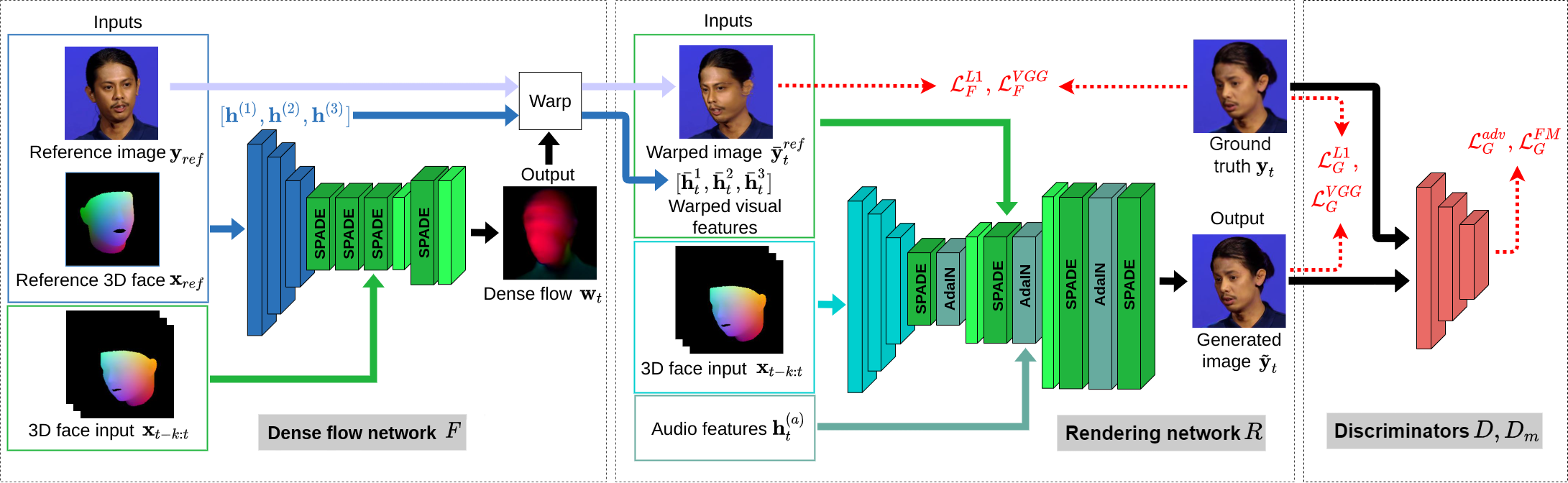}
\caption{Overview of \textit{HeadGAN}. The dense flow network $F$ computes a flow field for warping the reference image and features, according to the 3D face input. Then, the rendering network $R$ uses this visual information along with the audio features in order to translate the 3D face input into a photo-realistic image of the source.}
\label{fig:fig3}
\end{figure*}

\subsection{HeadGAN Framework}
\label{sec:headganframework}

Our GAN-based head reenactment system is equipped with a Generator driven by two modalities: 1) the \textit{3D face representation} extracted from the driving video and the reference image, 2) \textcolor{black}{optionally} the \textit{audio features} coming from the driver. Given $\textbf{x}_{t-k:t}$, the driving 3D face representation from frame $t$, concatenated channel-wise with the 3D faces coming from the past $k=2$ frames, the reference image $\textbf{y}_{ref}$ with the corresponding 3D face representation $\textbf{x}_{ref}$ and the audio feature vector $\textbf{h}^{(a)}_t$, our Generator hallucinates a photo-realistic image, given as
\begin{equation}
\label{eq:generator}
\tilde{\textbf{y}}_{t} = G(\textbf{x}_{t-k:t}, \textbf{y}_{ref}, \textbf{x}_{ref}, \textbf{h}^{(a)}_t; \theta_G).
\end{equation}
Conditioning synthesis on the spatio-temporal volume $\textbf{x}_{t-k:t}$ helps to achieve temporal coherence across frames. The reference image $\textbf{y}_{ref}$ provides information on the texture and appearance of the source person, while audio features enhance the generative ability of $G$ across the face, and mainly the mouth area. In more detail, the Generator consists of two sub-networks: \textit{a dense flow network} $F$ and a \textit{a rendering network} $R$. For an overview of the Generator $G$, please refer to Fig. \ref{fig:fig3}.

\noindent \textbf{Dense flow network $F$.} Our rendering network $R$ relies on high quality visual features that reflect the appearance of the source identity. Nonetheless, we observed that simply using an encoder to extract such features from the reference image $\textbf{y}_{ref}$, does not capitalise well on the potential of the rendering network's architecture. It has been proved more meaningful to align the visual feature maps with the desired head pose, which is reflected in the 3D face representation $\textbf{x}_{t}$, coming from the driving video. With this in mind, we propose a dense flow network, which learns a flow $\textbf{w}_t$ that can be used to warp visual features. For that, we pass the concatenation of the reference image and its corresponding 3D face $(\textbf{y}_{ref}, \textbf{x}_{ref})$ through an encoder, for the extraction of visual feature maps in three spatial scales $\textbf{h}^{(1)}, \textbf{h}^{(2)}, \textbf{h}^{(3)}$, which represent the appearance of the source identity. Then, a decoder predicts the flow $\textbf{w}_t$, guided by the driving 3D face representation $\textbf{x}_{t-k:t}$, which is injected into $F$ through SPADE blocks \cite{park2019SPADE}. Ideally, when applied on the reference image $\textbf{y}_{ref}$, this dense flow should yield a warped image of the source person, with the same head pose and expression, as shown in the driving 3D face representation $\textbf{x}_{t}$. By applying the flow field on each visual feature map, we obtain the warped visual features $\bar{\textbf{h}}^{(1)}_{t}, \bar{\textbf{h}}^{(2)}_{t}, \bar{\textbf{h}}^{(3)}_{t}$ and the warped reference image $\bar{\textbf{y}}_t^{ref}$, all of which depend on the driving head pose at frame $t$.

\noindent \textbf{Rendering network $R$.} At the core of our Generator, the rendering network aims to translate the 3D face representation $\textbf{x}_{t-k:t}$ into a photo-realistic image $\tilde{\textbf{y}}_{t}$ of the source. This is achieved with the assistance of high quality audio features $\textbf{h}^{(a)}_t$ and visual feature maps $\bar{\textbf{h}}^{(1)}_{t}, \bar{\textbf{h}}^{(2)}_{t}, \bar{\textbf{h}}^{(3)}_{t}$. First, an encoder receives $\textbf{x}_{t-k:t}$ as input and applies a sequence of convolutional layers with down-sampling. Then, a decoder consisting of alternating SPADE \cite{park2019SPADE} and AdaIN \cite{adain} layers generates the desired frame $\tilde{\textbf{y}}_{t}$. \textcolor{black}{These adaptive normalisation layers enable injecting 2D visual feature maps into the rendering network through SPADE blocks, as well as 1D audio features through AdaIN blocks.} As opposed to the original work on SPADE \cite{park2019SPADE}, where the conditional input of all SPADE layers is the same segmentation map down-sampled to match the spatial size of each layer, we capitalise on visual feature maps of multiple spatial scales $\bar{\textbf{h}}^{(1)}_{t}, \bar{\textbf{h}}^{(2)}_{t}, \bar{\textbf{h}}^{(3)}_{t}$, $\bar{\textbf{y}}_t^{ref}$ as modulation inputs to SPADE blocks. On the contrary, we pass the same audio feature vector $\textbf{h}^{(a)}_{t}$ to AdaIN blocks of all spatial scales. The decoder is further equipped with PixelShuffle layers \cite{pixelshuffle} for up-sampling, which contribute to the quality of generated samples.

\noindent \textbf{Discriminators $D$ and $D_m$.} The image Discriminator receives a synthetic pair $(\textbf{x}_{t}, \tilde{\textbf{y}}_{t})$, or a real one $(\textbf{x}_{t}, \textbf{y}_{t})$ and learns to distinguish between them. We use a second Discriminator $D_m$, which focuses on the mouth region. Apart from the real $\textbf{y}_{t}^{m}$ or generated $\tilde{\textbf{y}}_{t}^m$ cropped mouth area, this network is conditioned on the audio feature vector $\textbf{h}^{(a)}_{t}$, which is spatially replicated and then concatenated to the cropped images channel-wise.

\begin{table*}[bp]
\begin{center}
\begin{tabular}{|c||c|c|c|c|c|c||c|c|c|c|}
\hline
 & \multicolumn{6}{c||}{Reconstruction} & \multicolumn{4}{c|}{Reenactment} \\
\hline
Method & L1 $\downarrow$ & PSNR $\uparrow$ & LPIPS $\downarrow$ & FID $\downarrow$ & FVD $\downarrow$ & CSIM $\uparrow$ & FID $\downarrow$ & CSIM $\uparrow$ & ARD $\downarrow$ & AU-H $\downarrow$ \\
\hline\hline
\textit{X2Face}~\cite{X2Face} & 13.49 & 20.69 & 0.260 & 130.2 & 697 & 0.600 & 122.1 & 0.520 & 4.39 & 0.346 \\
\textit{fs-vid2vid}~\cite{fsvid2vid} & 17.15 & 18.52 & 0.197 & 62.8 & 471 & 0.542 & - & - & - & - \\
\textit{Bi-layer*}~\cite{bilayer} & \textit{12.18} & 20.19 & \textit{0.152} & 92.2 & 394 & 0.590 & 172.8 & 0.563 & \textbf{1.01} & \textbf{0.296} \\
\textit{FOMM}~\cite{FOMM} & 12.34 & \textit{20.93} & 0.153 & \textit{64.9} & \textit{338} & \textit{0.754} & \textit{63.7} & \textbf{0.765} & 12.53 & 0.400 \\
\textit{HeadGAN} & \textbf{11.32} & \textbf{21.46} & \textbf{0.112} & \textbf{36.1} & \textbf{254} & \textbf{0.807} & \textbf{58.0} & \textit{0.688} & \textit{1.35} & \textit{0.326} \\
\hline
\end{tabular}
\end{center}
\caption{Quantitative results on the tasks of reconstruction and reenactment for VoxCeleb \cite{voxceleb} test set. 
}
\label{table:rec_reen_table}
\end{table*}

\noindent \textcolor{black}{\textbf{Training Objective.} Networks \textit{F} and \textit{R} which constitute the Generator are optimised jointly. We train \textit{HeadGAN} on reconstruction, by applying perceptual and pixel losses $\mathcal{L}_F^{VGG}, \mathcal{L}_G^{VGG}$ and $\mathcal{L}_F^{L1}, \mathcal{L}_G^{L1}$, both on the warped and generated images, as seen in Fig.~\ref{fig:fig3} (red arrows). A GAN Hinge loss $\mathcal{L}_G^{adv}$ \cite{lim2017geometric} along with a feature matching loss $\mathcal{L}_G^{FM}$ \cite{xu2017learning} further increase the photo-realism of results. An extended discussion on the objective functions and networks architecture can be found in Appendices~\ref{appendix:B} and \ref{appendix:C} respectively.}

\subsection{Advantages of 3D face modeling}

The \textit{semantic information} enclosed within the 3D face representation allows our dense flow network $F$ to learn a precise flow of the facial region, as it provides a dense correspondence of facial points, between the reference and driving images. Compared to scene flow \cite{flownet3d, DeepFaceFlow} that can be obtained from 3D meshes, our flow field hallucinated by $F$ exists in areas where 3D representation is missing, such as hair and upper body, where warping is equally important. Furthermore, as 3DMMs allow to disentangle identity from expression, our choice to condition the rendering network $R$ on a 3D face extracted from the driving person and adapted to the identity characteristics of the source, enables \textit{HeadGAN} to tackle the identity preservation problem on the task of reenactment.

Lastly, modeling faces with 3DMMs makes \textit{HeadGAN} very efficient for video conferencing applications. It enables a \textit{"sender"} to efficiently compress a driving frame $\textbf{y}_t$, in the form of expression parameters $\textbf{p}_t^{exp} \in {\rm I\!R}^{28}$ and camera parameters $\textbf{c}_t \in {\rm I\!R}^{7}$, giving a total of 35 floating point values. Then, these parameters can be used by a \textit{"receiver"} to render the 3D face and use the Generator to reconstruct the driving frame. A single reference image needs to be sent once in the beginning of the session.

\section{Experiments}

\subsection{Implementation details}
\label{sec:details}

\noindent \textbf{3D Face Rendering.} Given a set of camera parameters $\textbf{c}$ and a 3D facial shape $\textbf{s} \in {\rm I\!R}^{3N}$ (see Eq. \ref{eq:3DMM}), we rasterise the 3D mesh and produce a visibility mask $I \in {\rm I\!R}^{H \times W}$ in the image plane. Each spatial location of $I$ stores the index of the corresponding visible triangle on the 3D face seen from this pixel. Then, we use the mean shape $\bar{\textbf{x}}$ of the 3DMM, in order to find the normalised x-y-z coordinates of the center of each visible triangle. In this way we obtain a 3D face representation $\textbf{x} \in {\rm I\!R}^{H \times W \times 3}$, where each pixel contains three coordinates. These values can be interpreted as colors and therefore texture of the 3D face \cite{PNCC}. 

\noindent \textbf{Dataset and Training.} We train and evaluate \textit{HeadGAN} on VoxCeleb \cite{voxceleb} dataset, which contains over 100,000 videos of 1,251 identities, at $256 \times 256$ resolution. We maintain the original train and test split. As a pre-processing step, we compute a 3D face image for each video frame in the dataset and extract per-frame audio feature vectors. During training, we perform self-reenactent, as we randomly sample the reference image from the target video. This provides access to ground truth data, enabling us to design reconstruction loss terms to train the Generator. For the optimisation of \textit{HeadGAN}, we use the ADAM \cite{adam} with $\beta_1 = 0.5$, $\beta_2 = 0.999$ and learning rate $\eta = 0.0002$, both for the Generator and Discriminator.

\begin{figure*}[h!]
\begin{picture}(285,285)
\put(15,22){\includegraphics[width=0.95\linewidth]{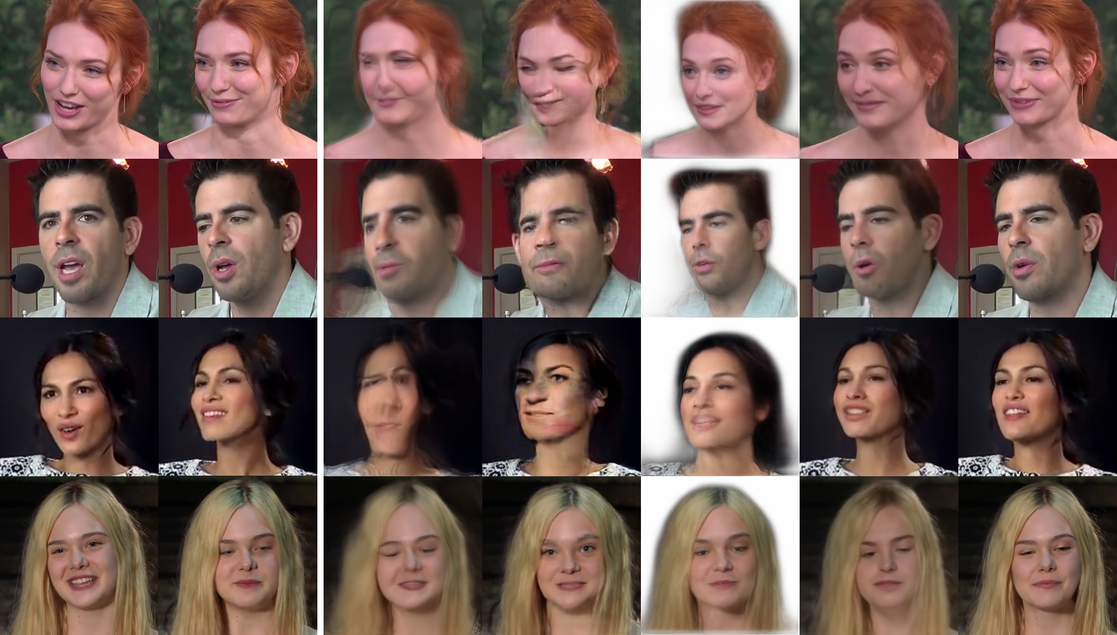}}
\put(30,10){Reference}
\put(100,10){Driving}
\put(162,10){\textit{X2Face}~\cite{X2Face}}
\put(220,10){\textit{fs-vid2vid}~\cite{fsvid2vid}}
\put(295,10){\textit{Bi-layer}~\cite{bilayer}}
\put(365,10){\textit{FOMM}~\cite{FOMM}}
\put(433,10){\textit{HeadGAN}}
\end{picture}
   \caption{Qualitative comparison with baselines, on the task of reconstruction (self-reenactment).}
\label{fig:reconstruction_figure}
\end{figure*}
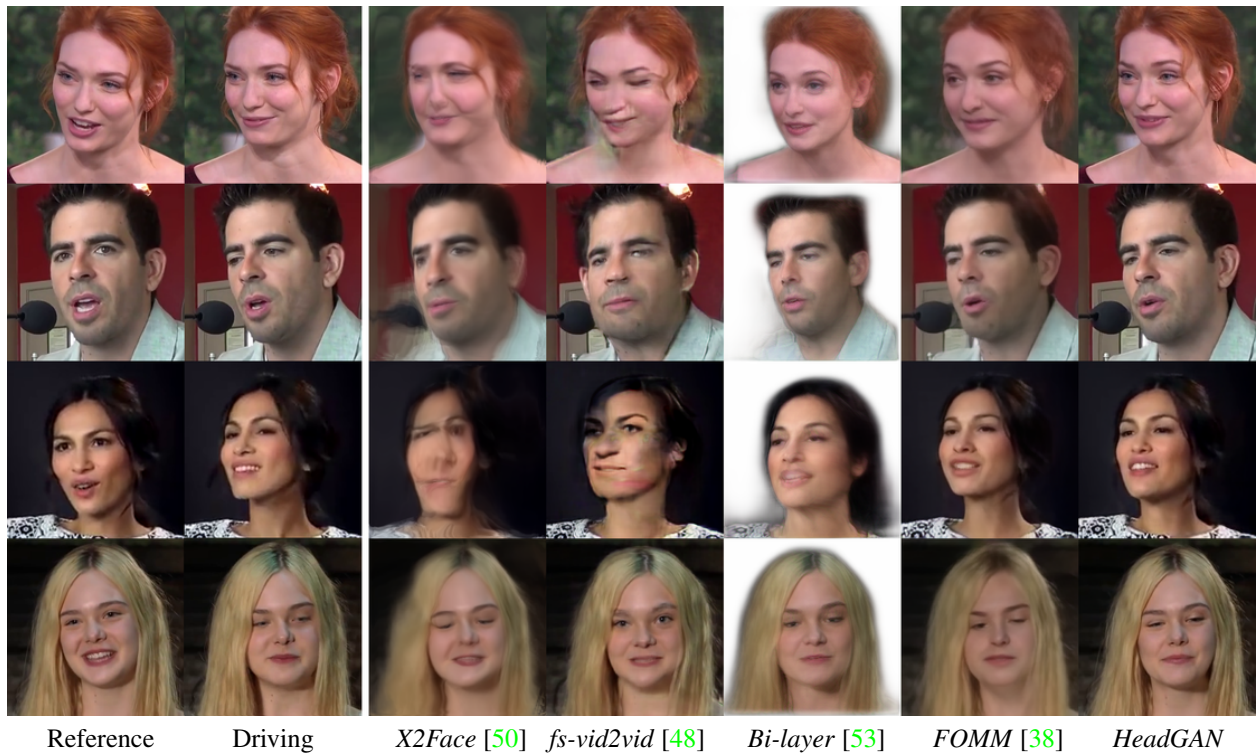

\subsection{Comparison with baselines}

\noindent \textbf{Reconstruction (self-reenactment).} First, we compare our approach with four state-of-the-art methods on the problem of self-reenactment, where the reference identity coincides with the driving one. Here, the task for \textit{HeadGAN} is to reconstruct the driving video from the 3D face representation sequence, using a single reference image to access appearance information. We perform both qualitative and quantitative comparisons with \textit{X2Face}~\cite{X2Face}, \textit{few-shot vid2vid}~\cite{fsvid2vid}, \textit{Bi-layer Neural Avatars}~\cite{bilayer} and \textit{First Order Motion Model}~\cite{FOMM}. For~\cite{X2Face} and~\cite{FOMM}, we use the pre-trained models provided by the authors, trained on VoxCeleb. We trained~\cite{fsvid2vid} from scratch, since no checkpoint was available. Lastly, we used the model provided by the authors of~\cite{bilayer}, trained on the larger VoxCeleb2* \cite{Chung18b}.

For the numerical evaluation of reconstruction we use L1 distance between the generated and ground truth frames, as well as Peak signal-to-noise ratio (PSNR) \textcolor{black}{and Learned Perceptual Image Patch Similarity (LPIPS) \cite{lpips}}. We access the realism of frames with \textcolor{black}{Fréchet Inception Distance} (FID) \cite{NIPS2017_8a1d6947} and \textcolor{black}{Fréchet Video Distance} (FVD) \cite{unterthiner2018towards} metrics. We use \textcolor{black}{Cosine Similarity} (CSIM)~\cite{deng2018arcface} to measure identity preservation. The results are presented in Table~\ref{table:rec_reen_table} and reveal that \textit{HeadGAN} outperforms all four baselines by a noteworthy margin, in every single metric.


The samples illustrated in Fig.~\ref{fig:reconstruction_figure} show that our method generates far more realistic images than the baselines with better preserved appearance traits.
We urge the reader to visually inspect the video results in our project's web page.

\begin{figure*}
\begin{picture}(285,285)
\put(15,22){\includegraphics[width=0.95\linewidth]{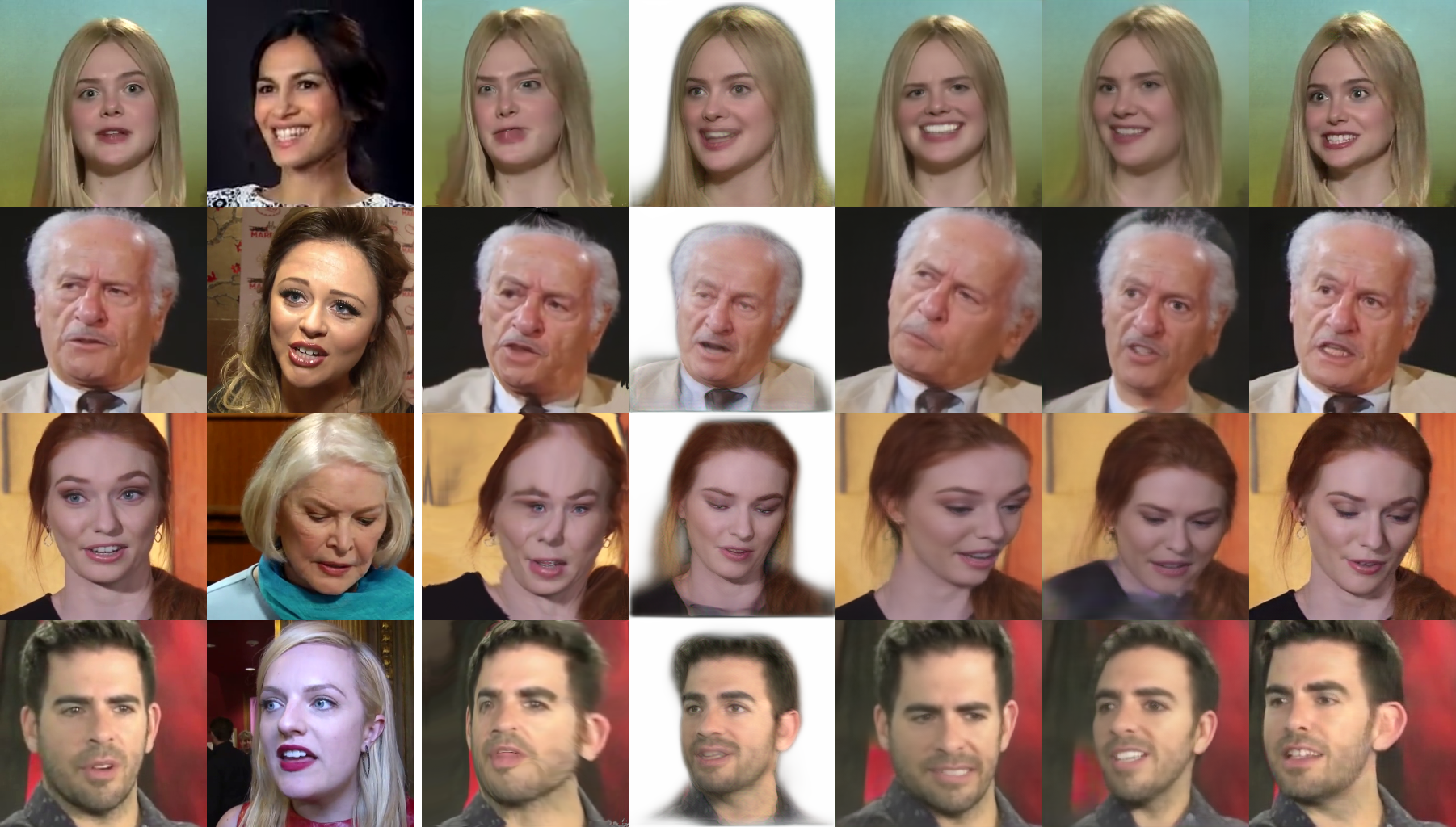}}
\put(30,10){Reference}
\put(100,10){Driving}
\put(162,10){\textit{X2Face}~\cite{X2Face}}
\put(227,10){\textit{Bi-layer}~\cite{bilayer}}
\put(286,10){\textit{FOMM-Rel}~\cite{FOMM}}
\put(354,10){\textit{FOMM-Abs}~\cite{FOMM}}
\put(433,10){\textit{HeadGAN}}
\end{picture}
   \caption{Qualitative comparison with baselines, on the task of reenactment. \textit{FOMM-Rel} refers to relative key-point coordinates strategy and \textit{FOMM-Abs} to absolute key-point coordinates, used in \textit{First Order Motion Model}~\cite{FOMM}.}
\label{fig:reenactment_figure}
\end{figure*}
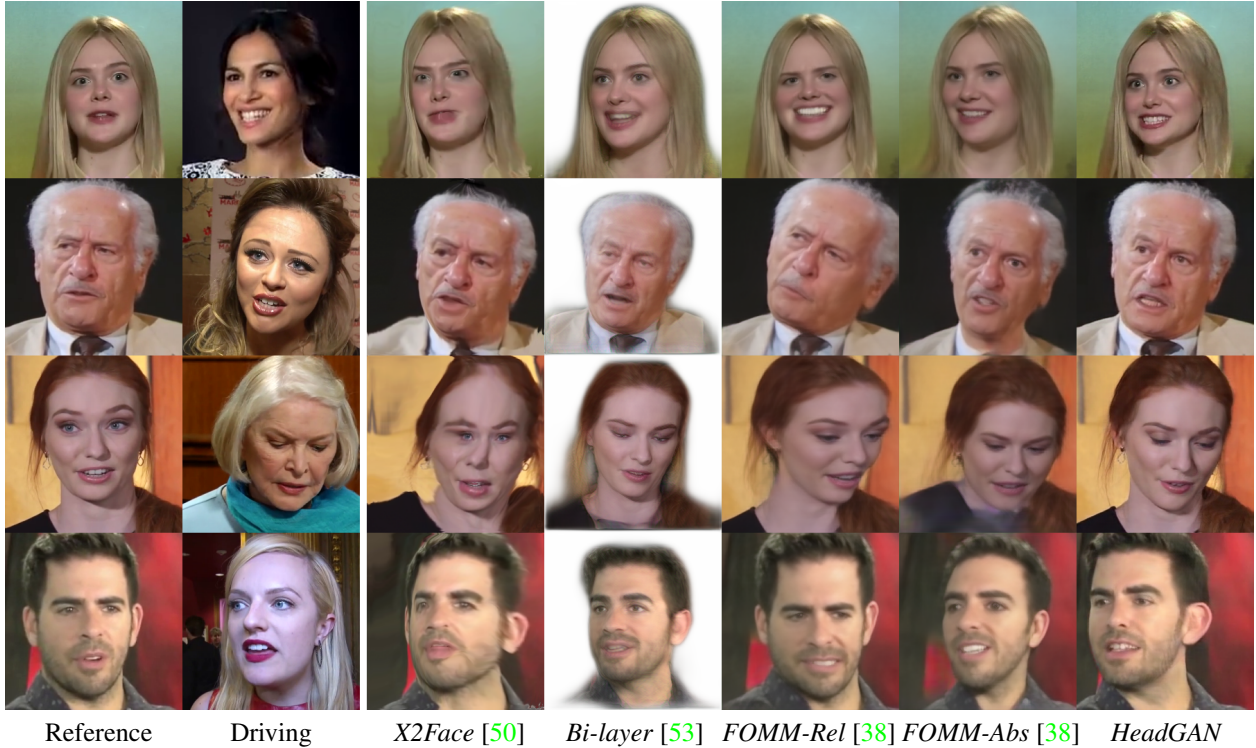

\noindent \textbf{Reenactment.} The objective of reenactment is to fully transfer the head pose and facial expressions of the target sequence to the person shown in the source image, while preserving the latter identity, as the driving and reference subjects are now different. To that end, we selected 15 random (video, image) pairs from VoxCeleb test set and performed reenactment, generating around 6K frames in total with each method. Apart from image quality (FID) and identity preservation (CSIM), we further evaluate the pose and expression transferability of systems, with Average Rotation Distance (ARD) in degrees and Action Units Hamming distance (AU-H) \cite{AUs}, respectively. Details on metrics can be found in Appendix~\ref{appendix:E}. As can be seen in Table~\ref{table:rec_reen_table}, \textit{HeadGAN} creates superior samples in terms of visual quality. \textit{Bi-layer Neural Avatars}~\cite{bilayer} performs slightly better on the task of facial expression transfer, which could be attributed to the fact that it was trained on the larger VoxCeleb2 \cite{Chung18b}. However, it performs poorly on identity preservation, as it conditions synthesis on facial landmarks, which unavoidably pass on identity related information from the driving subject. On the other hand, \textit{FOMM}~\cite{FOMM} uses relative key-point locations to account for the identity preservation problem that seems to increase CSIM. Nonetheless, this comes at the expense of pose transfer, as the model requires the face in the first frame of the driving video to have the same pose with the reference face, which is very rarely the case, also confirmed by the large ARD. When \textit{FOMM} uses absolute key-point locations instead, in order to accurately transfer pose, the identity preservation problem becomes apparent and CSIM drops to 0.587. This behaviour can be observed visually in Fig.~\ref{fig:reenactment_figure}, where the head geometry of the driver is reflected in the generated samples of \textit{FOMM-abs}. Different than the baselines, \textit{HeadGAN} performs well on all three requirements of successful reenactment (pose, expression transfer and identity preservation). Here we note that we have omitted comparisons with \textit{MarioNETte} \cite{MarioNETte:AAAI2020} and \textit{Warp-guided GANs}~\cite{Geng2018WarpguidedGF} since the source codes are not publicly available.

\begin{figure}[hb!]
\begin{picture}(120,120)
\put(0,22){\includegraphics[width=1.0\linewidth]{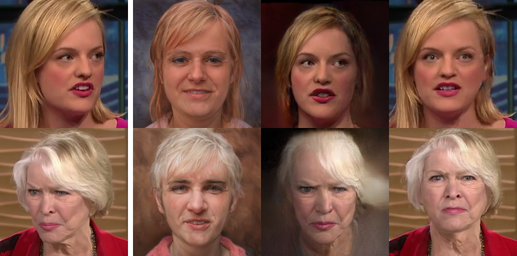}}
\put(5,10){Input image}
\put(75,10){\textit{psp}~\cite{psp}}
\put(131,10){\textit{RaR}~\cite{RaR}}
\put(186,10){\textit{HeadGAN}}
\end{picture}
   \caption{Qualitative comparison on frontalisation.}
\label{fig:frontalisation_figure}
\end{figure}
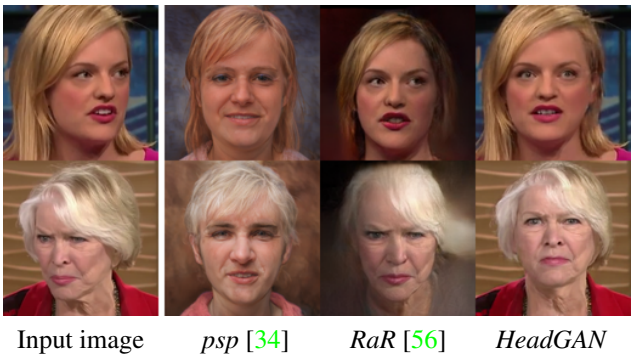

\begin{figure*}
\begin{picture}(160,160)
\put(0,22){\includegraphics[width=1.0\linewidth]{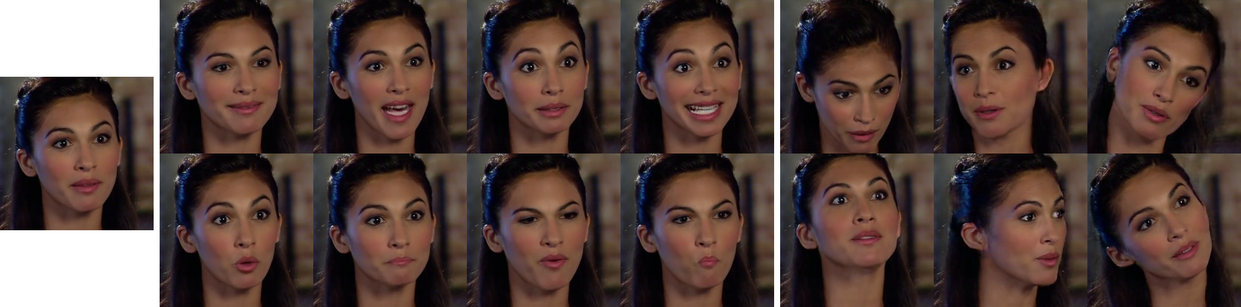}}
\put(5,40){Input image}
\put(60,150){Expression editing (top: positive value, bottom: negative value)}
\put(370,150){Rotation editing}
\put(78,10){$1^{st}$ p. c.}
\put(140,10){$2^{nd}$ p. c.}
\put(200,10){$3^{rd}$ p. c.}
\put(252,10){All three p. c.}
\put(333,10){Pitch}
\put(395,10){Yaw}
\put(455,10){Roll}
\end{picture}
   \caption{\textcolor{black}{Expression and Pose Editing. Please note that we use our model's variation without audio (AdaIN) for this task.}}
\label{fig:edit_figure}
\end{figure*}

\noindent \textbf{Frontalisation.} Our choice to use a 3D face representation to condition synthesis allows us to set the desired head pose manually, without the need of driving frames. By re-setting the camera parameters to frontal, we are able to generate a frontal view of the reference image. We compare \textit{HeadGAN} on the task of frontalisation with \textit{pixel2style2pixel (pSp)}~\cite{psp} and \textit{Rotate-and-Render (RaR)}~\cite{RaR}. For that, we randomly select one frame from each video of VoxCeleb test split and frontalise it. \textcolor{black}{Since there is no audio input for this task, we trained a variation of \textit{HeadGAN} without AdaIN layers for audio features.} Then, we measure the photo-realism of generated samples with FID, the identity preservation with CSIM and the Average Rotation Error (ARE) as the deviation from the frontal pose, in degrees. \textcolor{black}{We define frontal pose as zero Euler angles in camera rotation parameters coming from 3DMM fitting. Therefore, ARE is only used for reference here, as a sanity check.} The results are displayed in Table~\ref{table:frontalisation_table}. \textit{HeadGAN} performs equally well on CSIM with \textit{RaR} and surpasses baselines on image quality and frontalisation accuracy. Please see Fig. \ref{fig:frontalisation_figure} for a visual comparison.

\begin{table}[h!]
\begin{center}
\begin{tabular}{|c|c|c|c|}
\hline
Method & FID $\downarrow$ & CSIM $\uparrow$ & ARE $\downarrow$ \\
\hline\hline
\textit{psp}~\cite{psp} & 147.8 & 0.130 & 2.66 \\
\textit{RaR}~\cite{RaR} & \textit{88.4} & \textit{0.753} & \textit{2.65} \\
\textit{HeadGAN} & \textbf{30.1} & \textbf{0.766} & \textbf{0.76} \\
\hline
\end{tabular}
\end{center}
\caption{Quantitative results on frontalisation.}
\label{table:frontalisation_table}
\end{table}

\subsection{Image Expression and Pose Editing}

\textcolor{black}{Our model can be further used as an image editing tool.} Given a source image $\textbf{y}_{ref}$ and its shape and camera parameters $\textbf{p}_{ref}^{id}$, $\textbf{p}_{ref}^{exp}$ and $\textbf{c}_{ref}$, first we render the corresponding 3D face representation $\textbf{x}_{ref}$. Then, we re-adjust the expression or camera parameters manually and render a pseudo-driving 3D face $\textbf{x}_t$. We pass $\textbf{x}_t$, $\textbf{y}_{ref}$, $\textbf{x}_{ref}$ through the Generator to obtain a synthetic image with a novel expression or pose, reflecting the adjusted parameters. In Fig. \ref{fig:edit_figure} we show the results after manipulating the first three principal components (p. c.) of expression and the camera angles.

\subsection{Ablation Study}
\label{sec:ablation}
\begin{figure}[ht!]
\subfloat[Significance of dense flow network $F$ in image quality.]{%
  \includegraphics[scale=0.23]{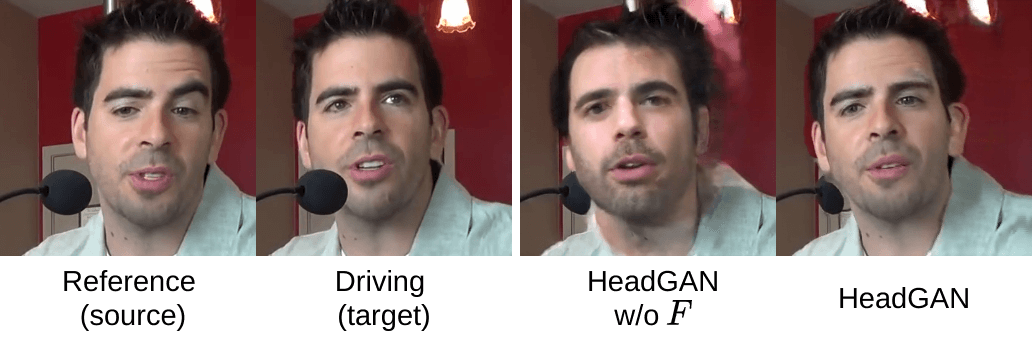}%
  \label{fig:fig5a}
}

\subfloat[The identity preservation problem becomes prominent when conditioning on facial landmarks, instead of the 3D face representation.]{%
  \includegraphics[scale=0.23]{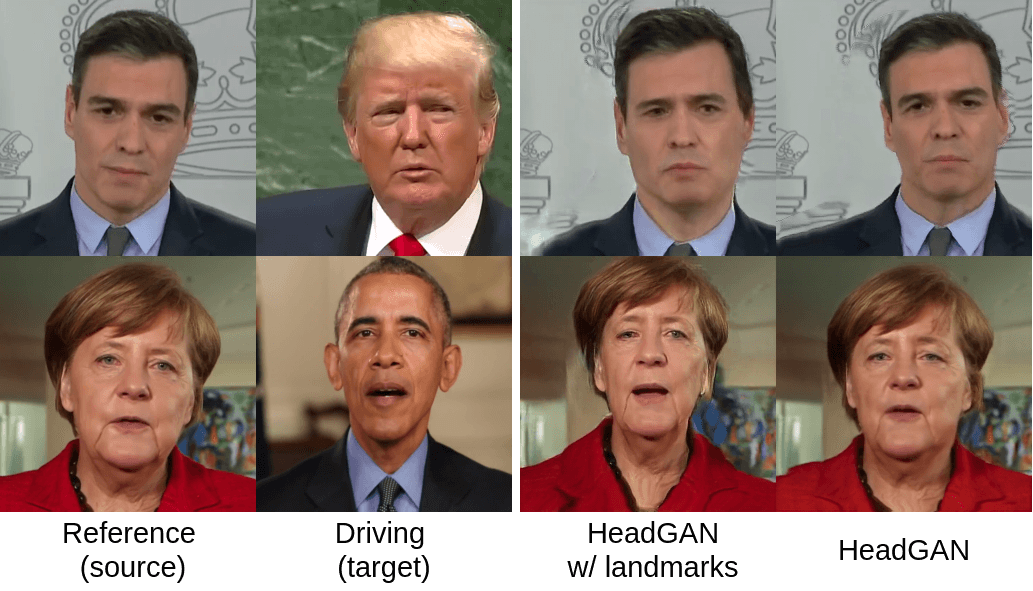}
  \label{fig:fig5b}
}

\label{fig:fig5}
\caption{The importance of \textit{HeadGAN} components.}
\end{figure}

We conducted an ablation study in order to assess 1) the significance of the dense flow network $F$, 2) the advantage of the 3D face representation \textcolor{black}{compared to sketches of 2D landmarks, that is the input used by \cite{fsvid2vid} and \cite{bilayer}}, 3) the contribution of audio modality. In order to evaluate the importance of $F$, we removed its decoding layers. We kept its encoder, which extracts appearance feature maps from the reference image. Then, instead of warping these features maps and reference image, we passed them directly to the SPADE layers of the rendering network $R$. In addition, we implemented a \textit{HeadGAN} variation with landmarks, by conditioning synthesis on sketch images, drawn by connecting 2D landmarks with edges \cite{fsvid2vid}. As can be seen in Table \ref{table:ablation_table}, the full model outperforms all variations. In terms of scores, we observe that flow network $F$ is an essential component of our system. The use of a 3D face representation (instead of landmarks) alleviates the identity preservation problem and can be better understood visually, in Fig. \ref{fig:fig5b}.

\begin{table}[h!]
\begin{center}
\begin{tabular}{|c|c|c|c|}
\hline
Method & FID $\downarrow$ & FVD $\downarrow$ & CSIM $\uparrow$ \\
\hline\hline
\textit{HeadGAN} w/o network $F$ & 63.3 & 473 & 0.307 \\
\textit{HeadGAN} w/ landmarks & 55.7 & 371 & 0.699 \\
\textit{HeadGAN} w/o audio input & 55.1 & 356 & 0.687 \\
\textit{HeadGAN} & \textbf{50.9} & \textbf{334} & \textbf{0.716} \\
\hline
\end{tabular}
\end{center}
\caption{Ablation study numerical results. Please note that we trained models for half epochs in this experiment.}
\label{table:ablation_table}
\end{table}

\noindent \textbf{Lipreading experiment.} We further evaluate the contribution of audio input to our system quantitatively, by employing an external lipreading network to classify synthetic videos. To that end, we chose 25 word classes of BBC dataset \cite{Chung16} and trained a lipreading classifier \cite{Stafylakis} on the default training split. After that, we reconstructed the test split of BBC dataset, using a random frame from the video as reference. We report a lipreading accuracy of $97\%$ on real test samples, $82\%$ on samples generated using the the full model and $73\%$ on synthetic data produced by the variation without considering audio input. These results suggest that the audio modality contributes largely on the generation of more plausible lip movements.

\section{Conclusion}

We presented \textit{HeadGAN}, a novel one-shot method for animating heads, driven by 3D facial data and audio features. Compared to SOTA methods, our framework exhibits superior reenactment performance and higher photo-realism. Our method can be further used for reconstruction, pose and facial expression editing, as well as frontalisation.

{\small
\bibliographystyle{ieee_fullname}
\bibliography{egbib}

\begin{thebibliography}{10}\itemsep=-1pt

\bibitem{openface}
Brandon Amos, Bartosz Ludwiczuk, and Mahadev Satyanarayanan.
\newblock Openface: A general-purpose face recognition library with mobile
  applications.
\newblock Technical report, CMU-CS-16-118, CMU School of Computer Science,
  2016.

\bibitem{elor2017bringingPortraits}
Hadar Averbuch-Elor, Daniel Cohen-Or, Johannes Kopf, and Michael~F. Cohen.
\newblock Bringing portraits to life.
\newblock {\em ACM Transactions on Graphics (Proceeding of SIGGRAPH Asia
  2017)}, 36(6):196, 2017.

\bibitem{AUs}
T. {Baltrušaitis}, M. {Mahmoud}, and P. {Robinson}.
\newblock Cross-dataset learning and person-specific normalisation for
  automatic action unit detection.
\newblock In {\em 2015 11th IEEE International Conference and Workshops on
  Automatic Face and Gesture Recognition (FG)}, volume~06, pages 1--6, 2015.

\bibitem{3dmms}
Volker Blanz and Thomas Vetter.
\newblock A morphable model for the synthesis of 3d faces.
\newblock In {\em Proceedings of the 26th Annual Conference on Computer
  Graphics and Interactive Techniques}, SIGGRAPH '99, page 187–194, USA,
  1999. ACM Press/Addison-Wesley Publishing Co.

\bibitem{Booth}
James Booth, Anastasios Roussos, Allan Ponniah, David Dunaway, and Stefanos
  Zafeiriou.
\newblock Large scale 3d morphable models.
\newblock {\em Int. J. Comput. Vision}, 126(2–4):233–254, Apr. 2018.

\bibitem{3dmmsbooth}
J. {Booth}, A. {Roussos}, E. {Ververas}, E. {Antonakos}, S. {Ploumpis}, Y.
  {Panagakis}, and S. {Zafeiriou}.
\newblock 3d reconstruction of “in-the-wild” faces in images and videos.
\newblock {\em IEEE Transactions on Pattern Analysis and Machine Intelligence},
  40(11):2638--2652, 2018.

\bibitem{Booth_2016_CVPR}
James Booth, Anastasios Roussos, Stefanos Zafeiriou, Allan Ponniah, and David
  Dunaway.
\newblock A 3d morphable model learnt from 10,000 faces.
\newblock In {\em Proceedings of the IEEE Conference on Computer Vision and
  Pattern Recognition (CVPR)}, June 2016.

\bibitem{chen2019hierarchical}
Lele Chen, Ross~K Maddox, Zhiyao Duan, and Chenliang Xu.
\newblock Hierarchical cross-modal talking face generation with dynamic
  pixel-wise loss.
\newblock In {\em Proceedings of the IEEE Conference on Computer Vision and
  Pattern Recognition}, pages 7832--7841, 2019.

\bibitem{Chung17b}
Joon~Son Chung, Amir Jamaludin, and Andrew Zisserman.
\newblock You said that?
\newblock In {\em British Machine Vision Conference}, 2017.

\bibitem{Chung18b}
J.~S. Chung, A. Nagrani, and A. Zisserman.
\newblock Voxceleb2: Deep speaker recognition.
\newblock In {\em INTERSPEECH}, 2018.

\bibitem{Chung16}
J.~S. Chung and A. Zisserman.
\newblock Lip reading in the wild.
\newblock In {\em Asian Conference on Computer Vision}, 2016.

\bibitem{deng2018arcface}
Jiankang Deng, Jia Guo, Xue Niannan, and Stefanos Zafeiriou.
\newblock Arcface: Additive angular margin loss for deep face recognition.
\newblock In {\em CVPR}, 2019.

\bibitem{RetinaFace}
Jiankang Deng, Jia Guo, Evangelos Ververas, Irene Kotsia, and Stefanos
  Zafeiriou.
\newblock Retinaface: Single-shot multi-level face localisation in the wild.
\newblock In {\em Proceedings of the IEEE/CVF Conference on Computer Vision and
  Pattern Recognition (CVPR)}, June 2020.

\bibitem{head2headpp}
Michail~Christos Doukas, Mohammad~Rami Koujan, Viktoriia Sharmanska, Anastasios
  Roussos, and Stefanos Zafeiriou.
\newblock Head2head++: Deep facial attributes re-targeting.
\newblock {\em IEEE Transactions on Biometrics, Behavior, and Identity
  Science}, 3(1):31--43, 2021.

\bibitem{Geng2018WarpguidedGF}
Jiahao Geng, T. Shao, Youyi Zheng, Y. Weng, and K. Zhou.
\newblock Warp-guided gans for single-photo facial animation.
\newblock {\em ACM Transactions on Graphics (TOG)}, 37:1 -- 12, 2018.

\bibitem{GIF2020}
Partha Ghosh, Pravir~Singh Gupta, Roy Uziel, Anurag Ranjan, Michael~J. Black,
  and Timo Bolkart.
\newblock {GIF}: Generative interpretable faces.
\newblock In {\em International Conference on 3D Vision (3DV)}, pages 868--878,
  2020.

\bibitem{giannakopoulos2015pyaudioanalysis}
Theodoros Giannakopoulos.
\newblock pyaudioanalysis: An open-source python library for audio signal
  analysis.
\newblock {\em PloS one}, 10(12), 2015.

\bibitem{MarioNETte:AAAI2020}
Sungjoo Ha, Martin Kersner, Beomsu Kim, Seokjun Seo, and Dongyoung Kim.
\newblock Marionette: Few-shot face reenactment preserving identity of unseen
  targets.
\newblock In {\em Proceedings of the AAAI Conference on Artificial
  Intelligence}, 2020.

\bibitem{deepspeech}
Awni Hannun, Carl Case, Jared Casper, Bryan Catanzaro, Greg Diamos, Erich
  Elsen, Ryan Prenger, Sanjeev Satheesh, Shubho Sengupta, Adam Coates, and
  Andrew Ng.
\newblock Deepspeech: Scaling up end-to-end speech recognition.
\newblock 12 2014.

\bibitem{NIPS2017_8a1d6947}
Martin Heusel, Hubert Ramsauer, Thomas Unterthiner, Bernhard Nessler, and Sepp
  Hochreiter.
\newblock Gans trained by a two time-scale update rule converge to a local nash
  equilibrium.
\newblock In I. Guyon, U.~V. Luxburg, S. Bengio, H. Wallach, R. Fergus, S.
  Vishwanathan, and R. Garnett, editors, {\em Advances in Neural Information
  Processing Systems}, volume~30, pages 6626--6637. Curran Associates, Inc.,
  2017.

\bibitem{adain}
X. {Huang} and S. {Belongie}.
\newblock Arbitrary style transfer in real-time with adaptive instance
  normalization.
\newblock In {\em 2017 IEEE International Conference on Computer Vision
  (ICCV)}, pages 1510--1519, 2017.

\bibitem{stylegan}
Tero Karras, Samuli Laine, and Timo Aila.
\newblock A style-based generator architecture for generative adversarial
  networks.
\newblock In {\em Proceedings of the IEEE/CVF Conference on Computer Vision and
  Pattern Recognition (CVPR)}, June 2019.

\bibitem{stylegan2}
Tero Karras, Samuli Laine, Miika Aittala, Janne Hellsten, Jaakko Lehtinen, and
  Timo Aila.
\newblock Analyzing and improving the image quality of {StyleGAN}.
\newblock In {\em Proc. CVPR}, 2020.

\bibitem{kim2018deep}
Hyeongwoo Kim, Pablo Garrido, Ayush Tewari, Weipeng Xu, Justus Thies, Matthias
  Nie{\ss}ner, Patrick P{\'e}rez, Christian Richardt, Michael Zoll{\"o}fer, and
  Christian Theobalt.
\newblock Deep video portraits.
\newblock {\em ACM Transactions on Graphics (TOG)}, 37(4):163, 2018.

\bibitem{adam}
Diederick~P Kingma and Jimmy Ba.
\newblock Adam: A method for stochastic optimization.
\newblock In {\em International Conference on Learning Representations (ICLR)},
  2015.

\bibitem{head2head2020}
M. Koujan, M. Doukas, A. Roussos, and S. Zafeiriou.
\newblock Head2head: Video-based neural head synthesis.
\newblock In {\em 2020 15th IEEE International Conference on Automatic Face and
  Gesture Recognition (FG 2020) (FG)}, pages 319--326, Los Alamitos, CA, USA,
  may 2020. IEEE Computer Society.

\bibitem{DeepFaceFlow}
M. Koujan, A. Roussos, and S. Zafeiriou.
\newblock Deepfaceflow: In-the-wild dense 3d facial motion estimation.
\newblock In {\em 2020 IEEE/CVF Conference on Computer Vision and Pattern
  Recognition (CVPR)}, pages 6617--6626, Los Alamitos, CA, USA, jun 2020. IEEE
  Computer Society.

\bibitem{vggloss}
C. {Ledig}, L. {Theis}, F. {Huszár}, J. {Caballero}, A. {Cunningham}, A.
  {Acosta}, A. {Aitken}, A. {Tejani}, J. {Totz}, Z. {Wang}, and W. {Shi}.
\newblock Photo-realistic single image super-resolution using a generative
  adversarial network.
\newblock In {\em 2017 IEEE Conference on Computer Vision and Pattern
  Recognition (CVPR)}, pages 105--114, 2017.

\bibitem{lim2017geometric}
Jae~Hyun Lim and Jong~Chul Ye.
\newblock Geometric gan, 2017.

\bibitem{flownet3d}
Xingyu Liu, Charles~R Qi, and Leonidas~J Guibas.
\newblock Flownet3d: Learning scene flow in 3d point clouds.
\newblock {\em CVPR}, 2019.

\bibitem{spectralnorm}
Takeru Miyato, Toshiki Kataoka, Masanori Koyama, and Yuichi Yoshida.
\newblock Spectral normalization for generative adversarial networks.
\newblock In {\em International Conference on Learning Representations (ICLR)},
  2018.

\bibitem{voxceleb}
A. Nagrani, J.~S. Chung, and A. Zisserman.
\newblock Voxceleb: a large-scale speaker identification dataset.
\newblock In {\em INTERSPEECH}, 2017.

\bibitem{park2019SPADE}
Taesung Park, Ming-Yu Liu, Ting-Chun Wang, and Jun-Yan Zhu.
\newblock Semantic image synthesis with spatially-adaptive normalization.
\newblock In {\em Proceedings of the IEEE Conference on Computer Vision and
  Pattern Recognition}, 2019.

\bibitem{psp}
Elad Richardson, Yuval Alaluf, Or Patashnik, Yotam Nitzan, Yaniv Azar, Stav
  Shapiro, and Daniel Cohen-Or.
\newblock Encoding in style: a stylegan encoder for image-to-image translation.
\newblock {\em arXiv preprint arXiv:2008.00951}, 2020.

\bibitem{Seitzer2020FID}
Maximilian Seitzer.
\newblock {pytorch-fid: FID Score for PyTorch}.
\newblock \url{https://github.com/mseitzer/pytorch-fid}, August 2020.
\newblock Version 0.1.1.

\bibitem{pixelshuffle}
W. {Shi}, J. {Caballero}, F. {Huszár}, J. {Totz}, A.~P. {Aitken}, R. {Bishop},
  D. {Rueckert}, and Z. {Wang}.
\newblock Real-time single image and video super-resolution using an efficient
  sub-pixel convolutional neural network.
\newblock In {\em 2016 IEEE Conference on Computer Vision and Pattern
  Recognition (CVPR)}, pages 1874--1883, 2016.

\bibitem{Siarohin_2019_CVPR}
Aliaksandr Siarohin, Stéphane Lathuilière, Sergey Tulyakov, Elisa Ricci, and
  Nicu Sebe.
\newblock Animating arbitrary objects via deep motion transfer.
\newblock In {\em The IEEE Conference on Computer Vision and Pattern
  Recognition (CVPR)}, June 2019.

\bibitem{FOMM}
Aliaksandr Siarohin, Stéphane Lathuilière, Sergey Tulyakov, Elisa Ricci, and
  Nicu Sebe.
\newblock First order motion model for image animation.
\newblock In {\em Conference on Neural Information Processing Systems
  (NeurIPS)}, December 2019.

\bibitem{ijcai2019-129}
Yang Song, Jingwen Zhu, Dawei Li, Andy Wang, and Hairong Qi.
\newblock Talking face generation by conditional recurrent adversarial network.
\newblock In {\em Proceedings of the Twenty-Eighth International Joint
  Conference on Artificial Intelligence, {IJCAI-19}}, pages 919--925.
  International Joint Conferences on Artificial Intelligence Organization, 7
  2019.

\bibitem{Stafylakis}
Themos Stafylakis and Georgios Tzimiropoulos.
\newblock Combining residual networks with lstms for lipreading.
\newblock {\em CoRR}, abs/1703.04105, 2017.

\bibitem{obama}
Supasorn Suwajanakorn, Steven Seitz, and Ira Kemelmacher.
\newblock Synthesizing obama: learning lip sync from audio.
\newblock {\em ACM Transactions on Graphics}, 36:1--13, 07 2017.

\bibitem{stylerig}
Ayush Tewari, Mohamed Elgharib, Gaurav Bharaj, Florian Bernard, Hans-Peter
  Seidel, Patrick P{\'e}rez, Michael Z{\"o}llhofer, and Christian Theobalt.
\newblock Stylerig: Rigging stylegan for 3d control over portrait images, cvpr
  2020.
\newblock In {\em {IEEE} Conference on Computer Vision and Pattern Recognition
  (CVPR)}. {IEEE}, june 2020.

\bibitem{thies2020nvp}
Justus Thies, Mohamed Elgharib, Ayush Tewari, Christian Theobalt, and Matthias
  Nie{\ss}ner.
\newblock Neural voice puppetry: Audio-driven facial reenactment.
\newblock {\em ECCV 2020}, 2020.

\bibitem{face2face}
J. Thies, M. Zollh{\"o}fer, M. Stamminger, C. Theobalt, and M. Nie{\ss}ner.
\newblock Face2face: Real-time face capture and reenactment of rgb videos.
\newblock In {\em Proc. Computer Vision and Pattern Recognition (CVPR), IEEE},
  2016.

\bibitem{instancenorm}
Dmitry Ulyanov, Andrea Vedaldi, and Victor~S. Lempitsky.
\newblock Instance normalization: The missing ingredient for fast stylization.
\newblock {\em CoRR}, abs/1607.08022, 2016.

\bibitem{unterthiner2018towards}
Thomas Unterthiner, Sjoerd van Steenkiste, Karol Kurach, Raphael Marinier,
  Marcin Michalski, and Sylvain Gelly.
\newblock Towards accurate generative models of video: A new metric \&
  challenges.
\newblock {\em arXiv preprint arXiv:1812.01717}, 2018.

\bibitem{Vougioukas}
Konstantinos Vougioukas, Stavros Petridis, and Maja Pantic.
\newblock Realistic speech-driven facial animation with gans.
\newblock {\em International Journal of Computer Vision}, 10 2019.

\bibitem{fsvid2vid}
Ting-Chun Wang, Ming-Yu Liu, Andrew Tao, Guilin Liu, Jan Kautz, and Bryan
  Catanzaro.
\newblock Few-shot video-to-video synthesis.
\newblock In {\em Conference on Neural Information Processing Systems
  (NeurIPS)}, 2019.

\bibitem{wang2018vid2vid}
Ting-Chun Wang, Ming-Yu Liu, Jun-Yan Zhu, Guilin Liu, Andrew Tao, Jan Kautz,
  and Bryan Catanzaro.
\newblock Video-to-video synthesis.
\newblock In {\em Advances in Neural Information Processing Systems (NeurIPS)},
  2018.

\bibitem{X2Face}
Olivia Wiles, A.~Sophia Koepke, and Andrew Zisserman.
\newblock X2face: A network for controlling face generation using images,
  audio, and pose codes.
\newblock In {\em Proceedings of the European Conference on Computer Vision
  (ECCV)}, September 2018.

\bibitem{xu2017learning}
Xiangyu Xu, Deqing Sun, Jinshan Pan, Yujin Zhang, Hanspeter Pfister, and
  Ming-Hsuan Yang.
\newblock Learning to super-resolve blurry face and text images.
\newblock In {\em Proceedings of the IEEE international conference on computer
  vision}, pages 251--260, 2017.

\bibitem{yang2016wider}
Shuo Yang, Ping Luo, Chen~Change Loy, and Xiaoou Tang.
\newblock Wider face: A face detection benchmark.
\newblock In {\em IEEE Conference on Computer Vision and Pattern Recognition
  (CVPR)}, 2016.

\bibitem{bilayer}
Egor Zakharov, Aleksei Ivakhnenko, Aliaksandra Shysheya, and Victor Lempitsky.
\newblock Fast bi-layer neural synthesis of one-shot realistic head avatars.
\newblock In {\em European Conference of Computer vision (ECCV)}, August 2020.

\bibitem{Zakharov2019FewShotAL}
E. Zakharov, Aliaksandra Shysheya, Egor Burkov, and V. Lempitsky.
\newblock Few-shot adversarial learning of realistic neural talking head
  models.
\newblock {\em 2019 IEEE/CVF International Conference on Computer Vision
  (ICCV)}, pages 9458--9467, 2019.

\bibitem{lpips}
{Richard Yi} Zhang, Phillip Isola, {Alexei A.} Efros, Eli Shechtman, and Oliver
  Wang.
\newblock The unreasonable effectiveness of deep features as a perceptual
  metric.
\newblock In {\em Proceedings - 2018 IEEE/CVF Conference on Computer Vision and
  Pattern Recognition, CVPR 2018}, Proceedings of the IEEE Computer Society
  Conference on Computer Vision and Pattern Recognition, pages 586--595. IEEE
  Computer Society, 2018.

\bibitem{RaR}
Hang Zhou, Jihao Liu, Ziwei Liu, Yu Liu, and Xiaogang Wang.
\newblock Rotate-and-render: Unsupervised photorealistic face rotation from
  single-view images.
\newblock In {\em IEEE Conference on Computer Vision and Pattern Recognition
  (CVPR)}, 2020.

\bibitem{PNCC}
X. Zhu, Z. Lei, X. Liu, H. Shi, and S.~Z. Li.
\newblock Face alignment across large poses: A 3d solution.
\newblock In {\em 2016 IEEE Conference on Computer Vision and Pattern
  Recognition (CVPR)}, pages 146--155, Los Alamitos, CA, USA, jun 2016. IEEE
  Computer Society.

\end{thebibliography}
}

\newpage
\phantom{a}
\newpage

\begin{appendices}

\section{3DMM fitting}
\label{appendix:A}

Given a facial image $\textbf{y}$, our 3DMM fitting stage recovers shape $\textbf{p}$ and camera $\textbf{c}$ parameters. It relies on dense 3D points of the face, which are regressed with RetinaFace-R501\cite{RetinaFace} network, pre-trained on WIDER FACE dataset \cite{yang2016wider}. We use Procrustes analysis to register the regressed points with the mean shape $\bar{\textbf{x}}$ of LSFM 3DMM \cite{Booth_2016_CVPR}.

\begin{algorithm}
\DontPrintSemicolon
\SetAlgoLined
\KwInput{3DMM: $\{ \textbf{U}^{id},  \textbf{U}^{exp}, \bar{\textbf{x}} \}$, image: $\textbf{y}$}

\tcc{Regress dense 3D points}
$\textbf{l}$ = RetinaFace($\textbf{y}$)

\tcc{Align points to mean shape}
$\textbf{l}'$ = Procrustes($\bar{\textbf{x}}, \textbf{l}$)

\tcc{Merge id. with exp. 3DMM}
$\textbf{U} = [ \textbf{U}^{id\top} ; \textbf{U}^{exp\top}]^{\top}$

\tcc{Compute Moore-Penrose inverse}
$\textbf{U}^{+} = (\textbf{U}^{\top} \textbf{U})^{-1} \textbf{U}^{\top} $

\tcc{Recover shape parameters}
$\textbf{p} = \textbf{U}^{+} (\textbf{l}' - \bar{\textbf{x}})$

\tcc{Compute affine camera matrix}
$\textbf{P} =$ Least\_squares ($\textbf{l}_{homog}, \bar{\textbf{x}}$) $\in {\rm I\!R}^{3 \times 4}$

\tcc{Recover camera parameters: scale, rotation, translation}
$\textbf{c} =$ P\_to\_srt($\textbf{P}$)

\KwResult{shape parameters $\textbf{p}$, camera parameters $\textbf{c}$}
\caption{Fit the 3DMM to a given image.}
\end{algorithm}

\section{Objective functions - Training}
\label{appendix:B}

We train HeadGAN framework, consisting of the Generator $G$ and the two Discriminators $D$ and $D_m$, using GAN Hinge loss \cite{lim2017geometric}. Therefore, the adversarial loss term for $G$ is given by
\begin{equation}
   \mathcal{L}_G^{adv} = - \mathrm{E}_{p_{data}}[D(\textbf{x}_{t}, \tilde{\textbf{y}}_{t}) + D_m(\textbf{h}^{(a)}_{t}, \tilde{\textbf{y}}_{t}^m)],
\end{equation}
where $\textbf{x}_{t}$ is the 3D face representation input, $\textbf{h}^{(a)}_{t}$ is the input audio feature vector, $\tilde{\textbf{y}}_{t}$ is the "fake" frame generated by $G$ and $\tilde{\textbf{y}}_{t}^m$ the corresponding cropped mouth area of size $64 \times 64$. Given that during training we perform self-reenactment, we have access to the ground truth frame $\textbf{y}_{t}$.
The image Discriminator $D$ is optimised by minimising the loss
\begin{equation}
\begin{split}
   \mathcal{L}_D^{adv} =  - \mathrm{E}_{p_{data}}[& \min(0, -1 + D(\textbf{x}_{t}, \textbf{y}_{t}) \\
   - & \min(0, -1 - D(\textbf{x}_{t}, \tilde{\textbf{y}}_{t})].
\end{split}
\end{equation}
and the mouth Discriminator $D_m$ using a similar loss
\begin{equation}
\begin{split}
   \mathcal{L}_{D_m}^{adv} =  - \mathrm{E}_{p_{data}}[& \min(0, -1 + D_m(\textbf{x}_{t}, \textbf{y}_{t}^m) \\
   - & \min(0, -1 - D_m(\textbf{x}_{t}, \tilde{\textbf{y}}_{t}^m)].
\end{split}
\end{equation}

The generative network $G$ is trained by minimising also a reconstruction loss term between the generated and ground frames, in the image pixel space
\begin{equation}
   \mathcal{L}_G^{L1} = \mathrm{E}_{p_{data}}[ || \tilde{\textbf{y}}_{t} - \textbf{y}_{t} ||_1],
\end{equation}
as well as the feature space, using feature maps extracted by a pre-trained VGG network \cite{vggloss}:
\begin{equation}
   \mathcal{L}_G^{VGG} = \mathrm{E}_{p_{data}}[\sum_l || VGG_l(\tilde{\textbf{y}}_{t}) - VGG_l(\textbf{y}_{t}) ||_1].
\end{equation}
Similarly to VGG loss, we use the two Discriminators to compute visual features from both real and synthetic frames and compute a feature matching loss $\mathcal{L}_G^{FM}$ that was originally proposed in \cite{xu2017learning} and has been proven very effective at increasing the photo-realism of generated samples.

In addition, we apply both $L1$ and VGG losses on the warped image $\bar{\textbf{y}}_t^{ref}$, in order to force the dense flow network $F$ to learn a correct flow from the reference image to the desired head pose, obtaining the loss terms $\mathcal{L}_F^{L1}$ and $\mathcal{L}_F^{VGG}$.

To sum up, the overall objective for $G$ is given as:
\begin{equation}
\begin{split}
   \mathcal{L}_G = & \mathcal{L}_G^{adv} + \lambda_{L1} \mathcal{L}_G^{L1} + \lambda_{VGG} \mathcal{L}_G^{VGG} + \lambda_{FM} \mathcal{L}_G^{FM} + \\
   & \lambda_{L1} \mathcal{L}_F^{L1} + \lambda_{VGG} \mathcal{L}_F^{VGG},
\end{split}
\end{equation}
with $\lambda_{L1} = 50$ and $\lambda_{VGG} = \lambda_{FM} = 10$. The Discriminators are optimised under their corresponding adversarial loss terms
\begin{equation}
\mathcal{L}_D = \mathcal{L}_D^{adv}, \quad \mathcal{L}_{D_m} = \mathcal{L}_{D_m}^{adv}.
\end{equation}

\section{Architecture Details}
\label{appendix:C}

\subsection{Generator $G$}

\noindent \textbf{Dense flow network} $F$ (Table \ref{table:1}). The dense flow network consists of an encoding and a decoding part. Its encoder is made up from three convolutional layers, each one with instance normalization units \cite{instancenorm} and ReLU activation functions. The last two convolutions are performed with a stride of 2, for down-sampling the input twice. The decoder is equipped with SPADE blocks \cite{park2019SPADE}, which are used to "inject" the 3D face representation $\textbf{x}_{t-k:t}$ (modulation input). Here we down-sample $\textbf{x}_{t-k:t}$ to match it with the spatial size of each SPADE layer, similarly to the original work \cite{park2019SPADE}. We employ two Pixel Shuffle \cite{pixelshuffle} layers, for up-sampling. Finally, dense flow is calculated with a $7\times7$ convolutional output layer.

\begin{table}[t!]
\centering
\begin{tabular}{c c c c}
\hline
& Block &  & Output size \\
\hline\hline
& Input &  & $(256, 256, 6)$ \\
$7 \times 7$ conv-32 & Inst. Norm. & ReLU & $(256, 256, 32)$ \\
$3 \times 3$ conv-128 & Inst. Norm. & ReLU & $(128, 128, 128)$ \\
$3 \times 3$ conv-512 & Inst. Norm. & ReLU & $(64, 64, 512)$ \\
\hline
& SPADE Block &  & $(64, 64, 512)$ \\
& SPADE Block &  & $(64, 64, 512)$ \\
& SPADE Block &  & $(64, 64, 512)$ \\
& Pixel Shuffle &  & $(128, 128, 128)$ \\
& SPADE Block &  & $(128, 128, 128)$ \\
& Pixel Shuffle &  & $(256, 256, 32)$ \\
\hline
& $7 \times 7$ conv-2  &  & $(256, 256, 2)$ \\
\hline
\end{tabular}
\caption{Architecture of dense flow network $F$.}
\label{table:1}
\end{table}

\begin{table}[t!]
\centering
\begin{tabular}{c c c c}
\hline
& Block &  & Output size \\
\hline\hline
& Input &  & $(256, 256, 9)$ \\
$7 \times 7$ conv-32 & Inst. Norm. & ReLU & $(256, 256, 32)$ \\
$3 \times 3$ conv-128 & Inst. Norm. & ReLU & $(128, 128, 128)$ \\
$3 \times 3$ conv-512 & Inst. Norm. & ReLU & $(64, 64, 512)$ \\
\hline
& SPADE Block &  & $(64, 64, 512)$ \\
& AdaIN Block &  & $(64, 64, 512)$ \\
& Pixel Shuffle &  & $(128, 128, 128)$ \\
& SPADE Block &  & $(128, 128, 128)$ \\
& AdaIN Block &  & $(128, 128, 128)$ \\
& Pixel Shuffle &  & $(256, 256, 32)$ \\
& SPADE Block &  & $(256, 256, 32)$ \\
& AdaIN Block &  & $(256, 256, 32)$ \\
& SPADE Block &  & $(256, 256, 32)$ \\
\hline
LReLU & $7 \times 7$ conv-3 & $\tanh$ & $(256, 256, 3)$ \\
\hline
\end{tabular}
\caption{Architecture of rendering network $R$.}
\label{table:2}
\end{table}

\noindent \textbf{Rendering network} $R$ (Table \ref{table:2}). Our rendering network has an encoder-decoder architecture as well. Its encoder has a similar structure to the encoder of $F$. The decoder is built from alternating SPADE and AdaIN blocks, which are used to condition synthesis on our multi-scale visual feature maps and audio feature vectors respectively. We use Pixel Shuffle layers for up-sampling, since we noticed it performs better than simple up-sampling operations (e.g. nearest neighbor, linear, bi-linear). After the last decoding block, a convolutional layer is placed for the computation of the synthetic RGB image.

\begin{figure}[t!]

\subfloat[SPADE Block architecture.]{%
  \includegraphics[scale=0.66]{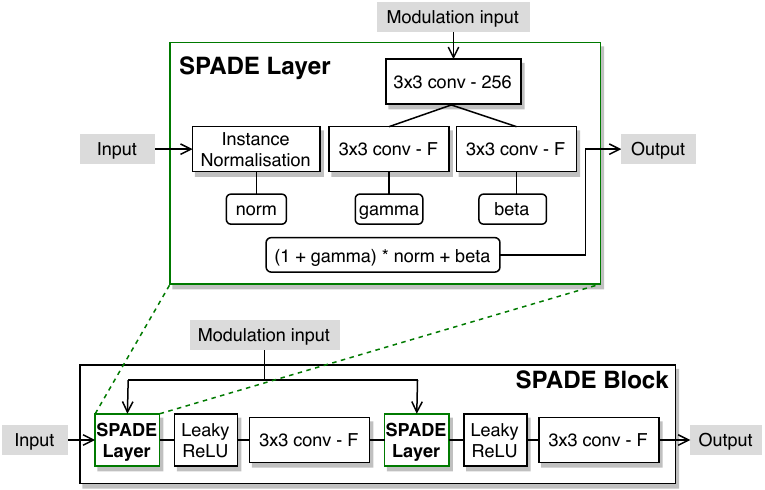}%
  \label{fig:fig1a}
}

\subfloat[AdaIN Block architecture]{%
  \includegraphics[scale=0.66]{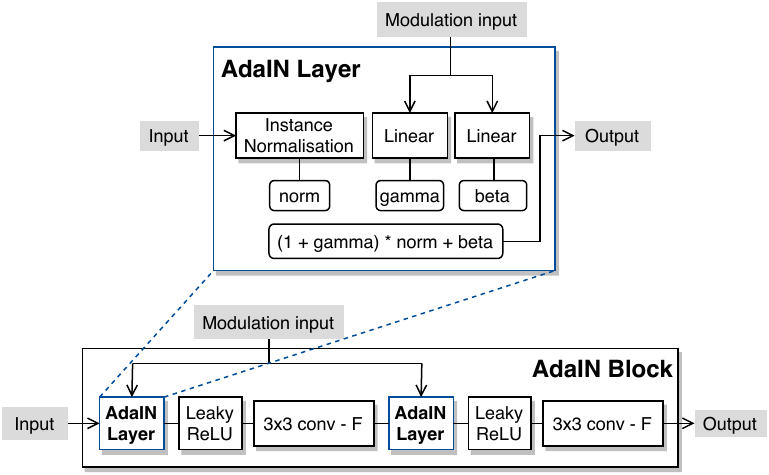}
  \label{fig:fig1b}
}

\label{fig:fig1}
\caption{Our SPADE and AdaIN blocks are based on the SPADE Resnet blocks proposed in \cite{park2019SPADE}, but without a residual component, as we always keep the same number of input channels F at the output, both on SPADE and AdaIN blocks.}

\end{figure}

\subsection{Discriminators $D$ and $D_m$}

Both $D$ and $D_m$ have a similar architecture to the discriminator presented in \cite{park2019SPADE}. We apply Spectral Normalisation \cite{spectralnorm} to all normalisation layers of the Discriminators.

\section{Additional results}

\begin{figure*}[h!]
\centering
\includegraphics[width=0.96\linewidth]{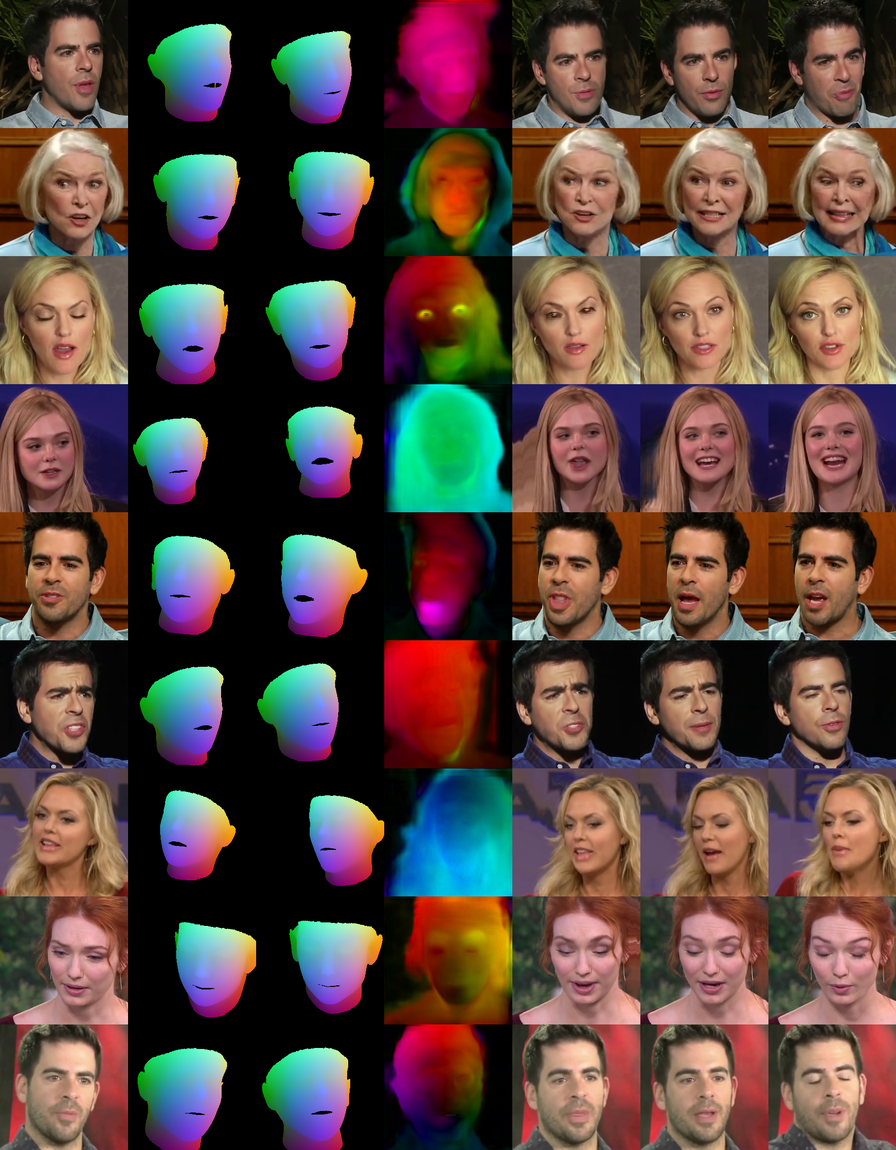}
\caption{Reconstruction. From left to right: reference, reference 3D face, driving 3D face, flow, warped, generated, driving.}
\label{fig:rec}
\end{figure*}

\begin{figure*}[h!]
\centering
\includegraphics[width=0.96\linewidth]{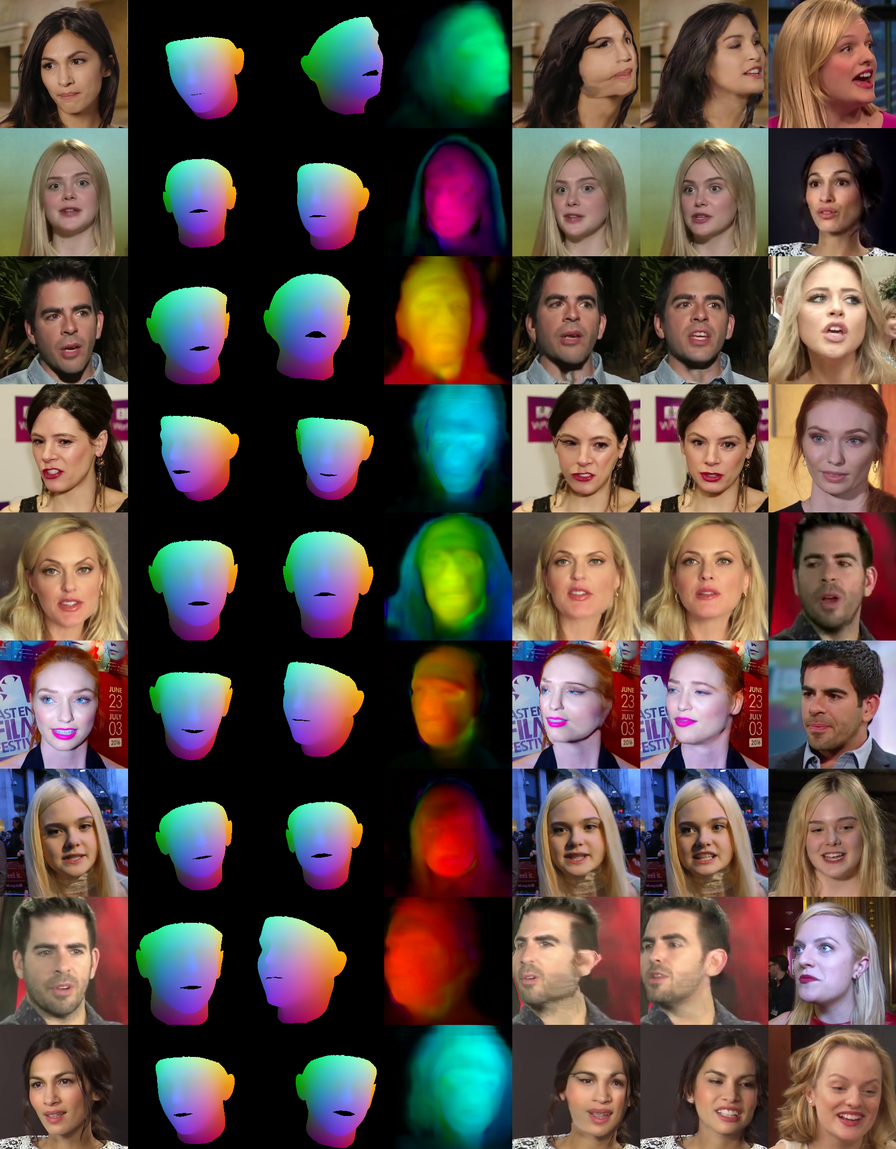}
\caption{Reenactment. From left to right: reference, reference 3D face, driving 3D face, flow, warped, generated, driving.}
\label{fig:ree}
\end{figure*}

In Fig. \ref{fig:rec} and Fig. \ref{fig:ree} we present a few more generated samples using VoxCeleb test set \cite{voxceleb}. Here, we also include the predicted flow and warped image in the results.

\section{Evaluation metrics}
\label{appendix:E}

We quantitatively compare \textit{HeadGAN} with the baselines, using the metrics described below.

\noindent \textbf{L1 distance (L1).} We evaluate the reconstructive ability of models by computing the mean $l1$-distance, between the synthesised and ground truth frames. We average the distance across channels, pixel locations and frames in the test set, to obtain the L1 metric. Please note that RGB channels are in the range [0, 255].

\noindent \textbf{Peak signal-to-noise ratio (PSNR).} This is another metric to measure the quality of reconstructed videos. PSNR is the ratio between the maximum possible power of a signal and the power of noise that affects the fidelity of its representation, defined as: $20 \cdot \log_{10} (\mathit{MAX}_I) - 10 \cdot \log_{10} \mathit{MSE}$. Here, $\mathit{MAX}_I=255$ and $\mathit{MSE}$ denotes the mean squared error, computed across color channels, spatial locations and frames. PSNR is expressed in dBs.

\noindent \textbf{Learned Perceptual Image Patch Similarity (LPIPS).} Perceptual metrics such as PSNR are simple shallow functions that are not able to account for many nuances of human perception. LPIPS \cite{lpips} uses a neural network that is trained to solve challenging visual prediction and modeling tasks as a feature extractor, since the network learns a representation that correlates well with perceptual judgments. Then, a similarity score between two images is calculated based on visual features.

\noindent \textbf{Fréchet Inception Distance (FID).} We employ FID \cite{NIPS2017_8a1d6947, Seitzer2020FID} as a measure of similarity between the dataset of real images and the dataset of images generated by the models. This score provides a useful insight into the photo-realism of synthetic frames.

\noindent \textbf{Fréchet Video Distance (FVD).} Given that we handle video data, it is important to evaluate the generative performance of models using a metric which takes into account the temporal coherence between frames. To that end, we calculate the FVD score \cite{unterthiner2018towards} of generated sequences, which has shown to correlate well with the human judgment on visual quality of generated videos.

\noindent \textbf{Cosine Similarity (CSIM).} Cosine similarity is a widely-used metric, which measures identity preservation in synthetic frames. We use ArcFace \cite{deng2018arcface} as an identity recognition network, in order to compute pairs of embedding vectors from the driving and corresponding generated images. Then, we calculate the cosine similarity between all pairs of embedding vectors in the dataset and report its average value. During reenactment, where no ground truth images are available, we extract the embedding vector from the reference image and compare it with the embeddings coming from synthetic frames. This leads to smaller CSIM values in reenactment, as the poses of the source and generated images do not match and the identity recognition network's output is not completely unaffected by a person's pose.

\noindent \textbf{Action Units Hamming distance (AU-H).} In order to measure the facial expression transferability of models, we use OpenFace \cite{openface} and more specifically \cite{AUs}, for the detection of Action Units (AU) in driving and generated images. Facial Action Coding System (FACS) is a system to taxonomise human facial movements by their appearance on the face. Using FACS it is possible to code nearly any anatomically possible facial expression, deconstructing it into the specific AUs that produced the expression. It is a common standard to objectively describe facial expressions. We use OpenFace to recognise if a set of AUs is present in a facial image or not, calculating a boolean vector of AUs. Then, we compute the Hamming distance $\in [0, 1]$ between AU boolean vectors, extracted from the corresponding driving and synthetic frames, and average across all frames.

\noindent \textbf{Average Rotation Distance (ARD).} This metric evaluates pose transfer. We use the camera parameters from 3D face reconstruction to compute the Euler angles that correspond to head poses in the driving and generated frames. Then, the average $l1$-distance of Euler angles across all frames is determined, in terms of degrees.

\noindent \textbf{Average Rotation Error (ARE).} This is a metric similar to ARD, but measures the average $l1$-distance of Euler angles from the frontal pose (zero degrees), across all images.

\end{appendices}

\end{document}